\documentclass{article}

\usepackage{microtype}
\usepackage{graphicx}
\usepackage{subfigure}
\usepackage{booktabs} 
\usepackage{amsfonts}
\usepackage{amsmath}
\usepackage{multirow}
\usepackage{tikz}
\usepackage{pgfplots}
\usetikzlibrary{bayesnet}

\usepackage{hyperref}

\usepackage{icml2020}

\newcommand{\pmh}[1]{}
\newcommand{\chl}[1]{}

\newcommand{\fig}[1]{Figure~\ref{fig:#1}}
\newcommand{\tab}[1]{Table~\ref{tab:#1}}
\newcommand{\sect}[1]{Section~\ref{sec:#1}}
\newcommand{\eqn}[1]{Eq.~\ref{eqn:#1}}

\newcommand{\given}{\, | \,}

\icmltitlerunning{Object-Centric Image Generation}

\begin{document}

\twocolumn[
\icmltitle{Object-Centric Image Generation with \\ Factored Depths, Locations, and Appearances}



\icmlsetsymbol{equal}{*}

\begin{icmlauthorlist}
\icmlauthor{Aeiau Zzzz}{equal,to}
\icmlauthor{Bauiu Yyyy}{equal,to,goo}
\icmlauthor{Cieua Vvvvv}{goo}
\end{icmlauthorlist}

\icmlaffiliation{to}{Department of Computation, University of Torontoland, Torontoland, Canada}
\icmlaffiliation{goo}{Googol ShallowMind, New London, Michigan, USA}

\icmlcorrespondingauthor{Cieua Vvvvv}{c.vvvvv@googol.com}
\icmlcorrespondingauthor{Eee Pppp}{ep@eden.co.uk}

\icmlkeywords{Machine Learning, ICML}

\vskip 0.3in
]



\printAffiliationsAndNotice{\icmlEqualContribution} 

\begin{abstract}
    We present a generative model of images that explicitly reasons over the set of objects they show.
    Our model learns a structured latent representation that separates objects from each other and from
    the background; unlike prior works, it explicitly represents the 2D position and depth of each object,
    as well as an embedding of its segmentation mask and appearance.
    The model can be trained from images alone in a purely unsupervised fashion without the need for object
    masks or depth information.
    Moreover, it always generates complete objects, even though a significant fraction of training images
    contain occlusions.
    Finally, we show that our model can infer decompositions of novel images into their constituent objects,
    including accurate prediction of depth ordering and segmentation of occluded parts.
\end{abstract}

\section{Introduction}
\label{sec:introduction}

As humans, we naturally understand the world in terms of \textit{objects}. 
We know the typical appearance of many types of object, and can imagine new instances of them---or indeed, entire scenes built from them.
Moreover, we can decompose a novel image in terms of the objects it shows, identifying their locations, distances and extents, and even reasoning about occluded parts. 
%

\begin{figure}
    \centering
    \includegraphics[width=\linewidth,trim=0 0 0 4,clip]{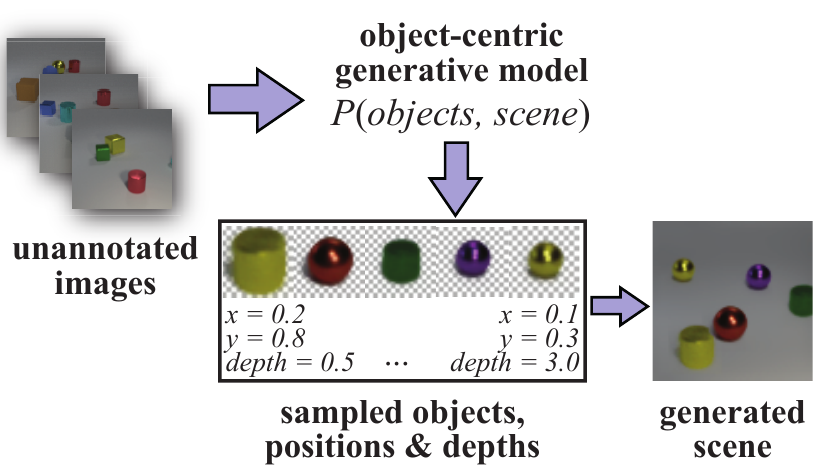}
    \vspace{-16pt}
    \caption{We present a probabilistic model of images, that learns to generate them by sampling a background and
    a set of foreground objects. Each object is associated with an appearance, mask, position, and depth.
    It is trained in a purely unsupervised fashion and learns to segment novel images, including occluded parts of objects}
    \label{fig:splash}
\end{figure}

Inspired by these human abilities, our goal in this work is to build a probabilistic generative model jointly over objects and entire scenes~(\fig{splash}).
We aim to learn this model from images showing scenes with several objects, often overlapping---yet like humans, to do so without relying on any annotations such as object masks or bounding boxes.
The final model should allow sampling images showing (i) complete, individual objects, and showing (ii) plausible scenes composed of them.
%

This learning task is highly challenging, as the model must simultaneously learn how to decompose images into their constituent objects, while also discovering their possible appearance variations.
We propose a way to solve this by embedding into the model knowledge about how images arise---the same object may appear at arbitrary positions; moreover, some objects are nearer than others, hence may hide parts of them.

We introduce these concepts into a generative model by using a \textit{structured} latent space and decoder (\sect{generative}).
The latent space has separate representations for each object and the background.
Each object has a classical, non-semantic,
appearance embedding to capture the details of its shape and texture, but also semantically-meaningful position and depth variables defining how it is placed in the image.
These latent variables are interpreted by a decoder that outputs each object independently, before assembling the final image according to their positions, alpha masks, and depths.
This final assembly process is not learnt, but rather leverages 
our prior knowledge about how images arise.
%

We see this model structure as a hybrid approach combining the best aspects of deep and classical generative models.
We incorporate our prior knowledge of how the world is constructed by including latent factors with well-defined meanings, but retain a black-box deep model for appearance variations that are difficult to characterize explicitly, such as irregular surfaces and illumination effects.

Our model is trained variationally to reconstruct images in terms of their constituent objects (\sect{train-inf}); thus, it learns to estimate object appearances, positions and depths, for previously-unseen images.
Notably, as it includes knowledge of the depth and the fact that objects may occlude one another when forming a 2D image, the model learns to predict the full (amodal) extent of substantially-occluded objects.

Some existing works have proposed generative models that include object structure. 
However, these are either (i) restricted to clearly-delimited objects over a plain background~\citep{eslami16nips,crawford19aaai}, or (ii) use full-image spatial mixture models that yield good segmentation results on complex scenes, but do not explicitly reason over object extents, locations and depths~\citep{greff19icml,burgess19arxiv,engelcke20iclr}.
In contrast, our model supports the generation of complex scenes, while still explicitly factoring out position and depth.

We demonstrate (\sect{experiments}) that our model learns to generate realistic images, achieving substantially better performance than a non-object-centric baseline of similar capacity.
These images are composed of objects which are themselves plausible in isolation. Furthermore, by manipulating the semantic dimensions of the latent representation we can influence the model to produce scenes with the characteristics we want.



To summarize, our contributions are:
\begin{itemize}
    \item the first method for image generation that explicitly factors out position and depth, allowing placement of objects with realistic occlusion effects, and avoiding the need for a deep encoder network to waste capacity modelling these,
    \item the first method able to reliably learn amodal segmentation of occluded objects without supervision,
    \item a novel approach to attentive placement and encoding of objects. 
\end{itemize}

\section{Related Work}
\label{sec:reld-work}


Deep generative models of images have received considerable attention following the seminal works on variational
autoencoders~\citep{kingma14iclr} and generative adversarial networks~\citep{goodfellow14nips}.
The vast majority of such models use a monolithic, unstructured latent space, that represents an entire scene in an unfactored fashion.
Recently, however, several generative models have emerged that use an object-centric representation instead.
%
\Citet{eslami16nips} present \textit{Attend, Infer, Repeat}, which is a VAE with recurrent encoder and decoder networks.
It learns to reconstruct images by attending to parts of them in sequence, controlled by an RNN, and placing generated patches back at the corresponding positions using a spatial transformer~\citep{jaderberg15nips}.
This model only supports images with black backgrounds and objects that never overlap.
It explicitly factors out the position in its latent space but does not consider depth nor amodal alpha masks.
%
\Citet{kosiorek18nips} extend this method to support video sequences, modelling object locations and presences over time.
The recent preprint of \citet{engelcke20iclr} presents a model that can generate much more complex scenes, including several objects over a background.
However, it uses a spatial mixture over objects; this means that there is no explicit factoring of position and depth, nor can the model produce amodal segmentations.
Instead, it must learn to model objects at different positions, with no guarantee that position is disentangled from appearance.

Other methods focus on unsupervised learning of discriminative tasks such as object detection and segmentation. 
The methods of \citet{greff16nips,greff17nips} learn to segment objects without supervision but do not reason over depth nor occlusions.
They cannot generate new images, nor have they been shown to scale to complex scenes.
\Citet{yuan19icml} present a method that does reason over occlusions but is restricted to objects of a single uniform color, overlaid on a known background.
The method of \citet{crawford19aaai} learns object detection without supervision, by auto-encoding images in terms of several objects placed using a grid of spatial transformers.
Similar to ours, this method incorporates a representation of object positions and depth; however, 
it is only shown on simple datasets, with uniform background colour and 2D sprites as objects.
Moreover, the authors do not attempt to evaluate segmentation, nor to generate new images.
%
\citet{greff19icml} and \citet{burgess19arxiv} present full-image spatial mixture models, trained with different iterative inference methods.
While technically generative, these last four methods cannot produce coherent scenes as they lack a scene-level latent representation~\citep{engelcke20iclr}.



\section{Generative Model}
\label{sec:generative}

\begin{figure}
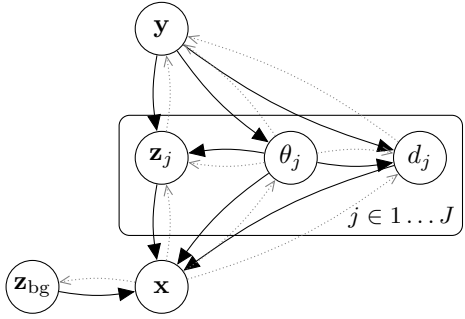

    \centering
    \tikz[]{
%
        \node[latent] (y) {$\mathbf{y}$} ;
        \node[latent, below=of y] (zj) {$\mathbf{z}_j$} ;
        \node[latent, right=of zj] (thj) {$\theta_j$} ;
        \node[latent, right=of thj] (dj) {$d_j$} ;
        \node[latent, below=of zj] (x) {$\mathbf{x}$} ;
        \node[latent, left=of x] (zbg) {$\mathbf{z}_\mathrm{bg}$} ;
        \plate{objects} {(zj) (thj) (dj)}  {$j \in 1 \ldots J$}
        \edge[bend right=10] {y} {zj, thj, dj} ;
        \edge[bend right=10,densely dotted,>=angle 45,color=gray] {zj, thj, dj} {y} ;
        \edge[bend right=10] {zbg, zj, thj, dj} {x} ;
        \edge[bend right=10,densely dotted,>=angle 45,color=gray] {x} {zbg, zj, thj, dj} ;
        \edge[bend right=10] {thj} {zj, dj} ;
        \edge[bend left=10,densely dotted,>=angle 45,color=gray] {thj} {zj, dj} ;
    }
    \caption{Our object-centric generative model of images, represented as a directed graphical model (belief network). Circles represent random variables; arrows represent conditional dependencies. Solid black arrows show dependencies in the generative model; dotted gray arrows show dependencies in the variational. See \sect{generative} for the meanings of the variable names}
    \label{fig:dgm}
\end{figure}

Our probabilistic model of images (\fig{dgm}) is a hierarchical Bayesian model 
with a first level of random variables modelling dependencies between different objects in the scene (e.g. placing objects such that they do not intersect), a second modelling dependencies among the pixels of each object, and a third modelling pixel-level noise.
%
%
We now describe this generative model in terms of the ancestral sampling process for an image $\mathbf{x}$ of size $N \times N$ pixels:
%
%
%
\begin{enumerate}
    \item sample a high-level embedding $\mathbf{y} \in \mathbb{R}^{D_\mathrm{scene}}$ for the scene, capturing dependencies among object appearances, positions and depths
    \item sample an appearance embedding $\mathbf{z}_\mathrm{bg} \in \mathbb{R}^{D_\mathrm{bg}}$ for the background, and decode it to pixels $\mathbf{x}_\mathrm{bg} \in (0 ,\, 1)^{N \times N}$
    \item for each object $j$, sample a position $\theta_j\in \{0 ,\, \ldots ,\,  N - 1\}^2$, conditioned on the scene embedding $\mathbf{y}$; this is an integer pixel coordinate indicating the center of the object in the image
    \item for each object $j$, sample an appearance embedding $\mathbf{z}_j \in \mathbb{R}^{D_\mathrm{obj}}$ and depth $d_j \in (0 ,\, 1)$; these are both conditioned on the positions of all objects, to account for (i) perspective effects (moving an object around in the image should change its appearance), and (ii) global illumination effects (objects may reflect or cast shadows on one another)
    \item independently decode the object appearance embeddings to pixels $\mathbf{x}_j \in (0 ,\, 1)^{M \times M}$ and alpha masks $\mathbf{\alpha_j} \in \{0 ,\, 1\}^{M \times M}$, where $M \leq N$ is the size of a `canvas' onto which the object is drawn at a canonical location, independent of where it will be placed in the final image 
    \item place the objects in an image according to their positions, depths, and alpha masks
    \item sample the final pixels $\mathbf{x} \in \mathbb{R}^{N \times N}$ from a Gaussian distribution centered on this mean image.
\end{enumerate}


In detail, we let
\begin{align}
    \mathbf{y}             &\sim  \mathcal{N}(\mathbf{0}, \mathbf{1}) \\
    \mathbf{z}_\mathrm{bg} &\sim  \mathcal{N}(\mathbf{0}, \mathbf{1}) \\
    \mathbf{x}_\mathrm{bg} &=     F_\mathrm{bg}(\mathbf{z}_\mathrm{bg}) \\
    \forall j \in \{1 ,\, \ldots ,\, J\}  \;\;
    \theta_j     &\sim  \mathrm{Categorical}(\,\cdot \given \zeta(\mathbf{y}))
\end{align}
\begin{equation}
    \forall j \in \{1 ,\, \ldots ,\, J\}
    \left\{
        \begin{aligned}
            z_j, d_j     &\sim  \mathcal{N}(\,\cdot \given \xi(\mathbf{y}, \, \theta_{1 \ldots J})) \\
            \mathbf{x}_j, \alpha_j  &=  F_\mathrm{obj}(\mathbf{z}_j) \\
        \end{aligned}
    \right.
\end{equation}
\begin{equation}
    \mathbf{x} \sim  \mathcal{N}(\,\cdot \given G(\mathbf{x}_\mathrm{bg} ,\, \mathbf{x}_{1 \ldots J} ,\, \alpha_{1 \ldots J} ,\, \theta_{1 \ldots J} ,\, d_{1 \ldots J}) ,\, \sigma^2)
\end{equation}
Here, $\xi$ is a densely-connected decoder network yielding the mean and variance of the object appearance priors.
$\zeta$ is a transpose-convolutional decoder yielding $N \times N$ values which we take as logits of a single categorical variable $\theta_j$, specifying the pixel coordinates of the object's center~\footnote{In practice we structure $\xi$ and $\zeta$ such that they operate on disjoint elements of $\mathbf{y}$; this allows us to intervene directly on the object positions $\theta_j$, i.e. to allow a user to directly manipulate the positions of objects in the scene}.
$F_\mathrm{bg}$ is a transpose-convolutional decoder mapping the background embedding to pixels, and $F_\mathrm{obj}$ is a transpose-convolutional decoder common to all objects mapping an object embedding to its pixels and alpha mask. 
$G$ is a \textit{compositor} function that combines the decoded objects and background according to the positions, depths, and masks, giving the final mean image (see below).
$\sigma$ is a constant specifying the magnitude of pixel noise.
Full details of the network architectures are given in the supplementary material.


The compositor function $G$ first places the decoded pixels $\mathbf{x}_j$ and alpha mask $\alpha_j$ of each object independently into separate, full-size ($N \times N$) images, by convolving each with a one-hot representation of its position:
\begin{align}
    \label{eqn:convl-placement}
    \mathbf{x}^*_j &= \mathbf{x}_j * \mathrm{one\text{-}hot}(\theta_j) \\
    \alpha^*_j     &= \alpha_j * \mathrm{one\text{-}hot}(\theta_j) \label{eqn:alpha-placement}
\end{align}
It then constructs the final image iteratively:
beginning with the background, and iterating objects from the farthest to the nearest, it alpha-blends each placed object in turn over the current image:
\begin{align}
    G(\cdot)      &= \mathbf{x}'_J \\
    \mathbf{s} &= \mathrm{argsort}_j \, d_j^{-1} \\
    \mathbf{x}'_j &= (1 - \alpha^*_{s_j}) \, \mathbf{x}'_{j - 1} + \alpha^*_{s_j} \mathbf{x}^*_{s_j} \\
    \mathbf{x}'_0 &= \mathbf{x}_\mathrm{bg}
\end{align}

%


%
%
%
%

%
%

\section{Training and Inference}
\label{sec:train-inf}

We use amortized variational inference~(AVI)~\citep{rezende14icml,kingma14iclr}, introducing a variational posterior distribution parametrized by an encoder network.
The encoder network must predict parameters of the posterior distribution over latent variables `explaining' a given image; as our latent space represents objects separately, the encoder learns to decompose an image into its constituent objects.
The encoder and generative model are then trained jointly by stochastic gradient descent.

\paragraph{Continuous relaxation.}
To train with gradient descent, the generative process must be fully differentiable.
However, the generative model of \sect{generative} includes discrete variables (the binary alpha masks and categorical positions), and $G$ is itself discontinuous due to the dependence on the ordering of the depths $d_j$.
We therefore relax the model, by
(i) changing the binary masks $\alpha_j$ to be continuous variables taking values in $(0 ,\, 1)$;
(ii) replacing the Categorical distribution on positions $\theta_j$ with a Gumbel-Softmax distribution~\citep{maddison17iclr,jang17iclr}, so $\theta_j$ defines continuous weights for all pixels, instead of selecting just one;
and (iii) replacing the hard depth ordering in $G$ with the heuristic softened version proposed by \citet{liu19iccv}.

\paragraph{Variational posterior.}
%
We use a structured variational distribution $Q$, factorized as
\begin{multline}
    Q(\mathbf{y} \given \mathbf{z}_{1 \ldots J} ,\, \theta_{1 \ldots J} ,\, d_{1 \ldots J}) \times
    Q(\mathbf{z}_\mathrm{bg} \given \mathbf{x}) \times \\ 
    \prod_{j=1}^J Q(\mathbf{z}_j \given \theta_j ,\, \mathbf{x}) \, Q(d_j \given \theta_j ,\, \mathbf{x}) \, Q(\theta_j \given \mathbf{x})
\end{multline}
%
%
This factorization is shown by the dotted gray lines in \fig{dgm}.
For $Q(\theta_j \given \mathbf{x})$ we use a Categorical distribution, with logits for all objects produced by a U-Net~\citep{ronneberger15miccai} architecture.
For the remaining terms, we use diagonal-covariance Gaussian distributions, parametrized by a convolutional network.
Details of the network architectures are given in the supplementary material.
%

In practice we achieved good results by separating the training of the scene-level and object-level parts of the model.
We first train $F_\mathrm{obj}$ and $F_\mathrm{bg}$ jointly with the encoders for $\mathbf{z}_j$, $\theta_j$ and $d_j$, 
treating the objects as fully independent of one another, with $\zeta$ taken to be uniform and $\xi$ standard Gaussian.
Then, we train $\zeta$ and $\xi$ jointly with the encoder for $\mathbf{y}$, to match the modes of the variational posteriors on $\mathbf{z}_j$, $\theta_j$ and $d_j$.
%
%

\paragraph{Attentive encoding.}
We find that it is beneficial for some datasets to incorporate a spatial attention mechanism in the encoder networks for $\mathbf{z}_j$ and $d_j$.
%
This
extracts areas of the image centered on locations where $\theta_j$ assigns large weight.
Specifically, we calculate the 2D cross-correlation of the image with $\theta_j$; this is equivalent to extracting a crop at every possible pixel location, and averaging these weighted by the corresponding values in $\theta_j$.
This process precisely mirrors the placement of objects in the image by convolution in the generative (\eqn{convl-placement})
\footnote{These large convolution and cross-correlation operations can be evaluated efficiently in Fourier space}.
The extracted image region is passed to a CNN predicting $\mathbf{z}_j$ and $d_j$.
%

\newcommand{\genrow}[2]{ 
    \includegraphics[scale=0.4]{gen-images/#1/#2_final.png} &
    \includegraphics[scale=0.4]{gen-images/#1/#2_segmentation.png} &
    \includegraphics[scale=0.6]{gen-images/#1/#2_obj_0.png} &
    \includegraphics[scale=0.6]{gen-images/#1/#2_obj_1.png} &
    \includegraphics[scale=0.6]{gen-images/#1/#2_obj_2.png} &
    \includegraphics[scale=0.6]{gen-images/#1/#2_obj_3.png} &
    \includegraphics[scale=0.6]{gen-images/#1/#2_obj_4.png} &
    \includegraphics[scale=0.4]{gen-images/#1/#2_bg.png} &
    \includegraphics[scale=0.4]{gen-images/#1/#2_locations.png}  \\
}

\newcommand{\genrowthree}[2]{ 
	\includegraphics[scale=0.4]{gen-images/#1/#2_final.png} &
	\includegraphics[scale=0.4]{gen-images/#1/#2_segmentation.png} &
	\includegraphics[scale=0.5]{gen-images/#1/#2_obj_0.png} &
	\includegraphics[scale=0.5]{gen-images/#1/#2_obj_1.png} &
	\includegraphics[scale=0.5]{gen-images/#1/#2_obj_2.png} &
	\includegraphics[scale=0.5]{gen-images/#1/#2_obj_3.png} &
	\includegraphics[scale=0.5]{gen-images/#1/#2_obj_4.png} &
	\includegraphics[scale=0.4]{gen-images/#1/#2_bg.png} &
	\includegraphics[scale=0.4]{gen-images/#1/#2_locations.png}  \\
}

\newcommand{\genrowfive}[2]{ 
	\includegraphics[scale=0.4]{gen-images/#1/#2_final.png} &
	\includegraphics[scale=0.4]{gen-images/#1/#2_segmentation.png} &
	\includegraphics[scale=0.6]{gen-images/#1/#2_obj_0.png} &
	\includegraphics[scale=0.6]{gen-images/#1/#2_obj_1.png} &
	\includegraphics[scale=0.6]{gen-images/#1/#2_obj_2.png} &
	\includegraphics[scale=0.6]{gen-images/#1/#2_obj_3.png} &
	\includegraphics[scale=0.6]{gen-images/#1/#2_obj_4.png} &
	\includegraphics[scale=0.4]{gen-images/#1/#2_bg.png} &
	\includegraphics[scale=0.4]{gen-images/#1/#2_locations.png}  \\
}

\begin{figure*}[t!]
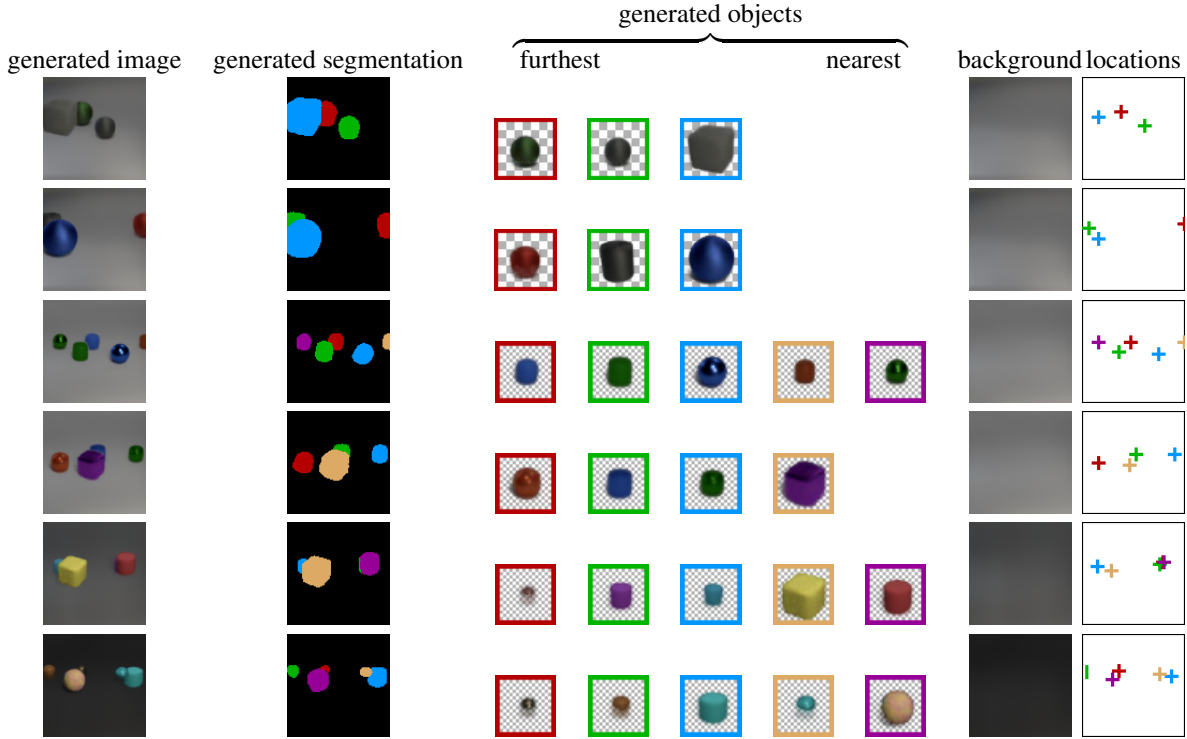

    \centering
    \begin{tabular}{@{}cccccccc@{}c}
        generated image & generated segmentation & \multicolumn{5}{c}{
            $\overbrace{\text{~furthest}\hspace{3cm}\text{nearest~}}^{\mbox{\normalsize\text{generated objects}}}$
        } & background & locations \\
        \genrowthree{0}{499}
        \genrowthree{0}{035}
        \genrowfive{1}{016}
        \genrowfive{1}{277}
        \genrowfive{7}{339}
        \genrowfive{7}{517}
    \end{tabular}
    \\
    \caption{Examples of objects and images generated by our model, on CLEVR-3 (top two rows), CLEVR-5 (middle two rows), and CLEVR-5-vbg (bottom two rows). Each row shows a final image, its modal segmentation into objects (colors) and background (black), and the objects and background from which it is assembled; we omit objects with all alpha values less than $0.5$ \pmh{?}. The objects are ordered by increasing depth from left to right; each is alpha-blended over a checkerboard pattern to clearly show its extent.}
    \label{fig:gen-examples}
\end{figure*}

\paragraph{Training objective.}
We maximize the evidence lower bound (ELBO)~\cite{blei17jasa,kingma14iclr} with respect to all network parameters, but modify the weight given to the KL-divergence following \citet{higgins17iclr}.
In order to encourage use of the factored position variable, rather than varying the placement of the object within its own canvas $(\mathbf{x}_j ,\, \alpha_j)$, we replace the KL term for $\theta_j$ with the $L^1$ divergence between the aggregated posterior $\mathbb{E}_\mathbf{x} \big\{ \frac{1}{J} \sum_j Q(\theta_j \given \mathbf{x}) \big\}$ and the uniform prior. 
%
A specific property of the way we render the object to its location
(as convolution with a related discrete location indicator) is that 
we obtain useful gradients regardless of the object position. 
This is contrast to models based on spatial-transformers, such as
AIR, which can get stuck in local minima if a good guess of the 
object locations is not available. 

Further details and hyperparameters are given in the supplementary material.



\section{Experiments}
\label{sec:experiments}
In this section, we demonstrate that our model is able to sample realistic images composed of multiple objects, by generating a set of objects, each of which is plausible in isolation, and arranging them coherently in the image.
We also show that the encoder network successfully learns to decompose novel images into their constituent objects, disentangling position from appearance and depth, and estimating segmentation masks.
Finally, we highlight the advantages of a factorized latent
space with semantic components by illustrating the effect of
interpolating between images via their latent codes, varying
either only their appearance, or only the objects' position,
or both.

We emphasize that in all experiments, our model achieves these results while learning in a fully-unsupervised fashion, using only unannotated images as input.
Full details of hyperparameter tuning and training infrastructure are given in the supplementary material.

\paragraph{Datasets.}
We train and evaluate our approach on three datasets of images---\textit{multi-dSprites}, \textit{overlapping-polygons}, and \textit{CLEVR}\footnote{Interestingly, we found that the effect of spatial attention in the encoder differed between datasets: for multi-dSprites it was highly beneficial, so we used it in all experiments; conversely, for CLEVR, it hindered convergence, while it had no significant effect for overlapping-polygons. We are not aware that such an effect has been reported in the literature before, and we plan to explore the reasons in future work.}.
Each image in multi-dSprites~\citep{burgess19arxiv} shows a 2D arrangement of shapes such as hearts and squares, with randomly-sampled positions, rotations, and colors, overlaid from back to front in a random order.
For our experiments, we use the original dataset, but take only the first 100K images containing 2--4 objects.
Each image in overlapping-polygons shows two 2D polygons with 3--6 sides, and randomly sampled colors, radii, and positions.
We include this dataset as its images have a higher degree of overlap than the others, thus validating our model in a more challenging setting.
Each image in CLEVR shows a 3D arrangement of several objects such as cubes and cylinders, with randomly-sampled positions and material properties; we generated these using a modified version of the code from \citet{johnson17cvpr}.
We use three variants with have complementary characteristics:
CLEVR-3 has 2--3 objects per image, with a moderately high degree of overlap;
CLEVR-5 has 3--5 slightly smaller objects per image with lower average overlap;
CLEVR-5-vbg is similar but the brightness (diffuse reflectivity) of the background is sampled randomly.
%
As these datasets consist of synthetic images, it is possible to generate perfect ground-truth annotations (including segmentations of occluded parts), allowing accurate benchmarking of our model's performance on segmentation and depth estimation.
%
%
Further details about all the datasets, train/validation splits, etc. are given in the supplementary material.
The datasets and generation code will be made public soon.

\begin{table}
    \centering
    \caption{Quantitative generation results: we report the FID and KID~$\times 100$ both for individual objects and for full scenes. We also include baseline values from an unstructured convolutional VAE; note that our method substantially outperforms this baseline, in spite of it having similar capacity and architecture. For both metrics, lower values are better. \pmh{update CLEVR-3 CVAE numbers}}
    \label{tab:gen-results}
    \vspace{4pt}
    \begin{tabular}{@{}l @{~~}c@{~}c @{~~}c@{~}c @{~~}c@{~}c@{}}
        \toprule
         & \multicolumn{2}{@{~~}c}{\textit{per-object}} & \multicolumn{2}{@{~~}c}{\textit{full-scene}} & \multicolumn{2}{@{~~}c@{}}{\textit{VAE baseline}} \\
         & FID & KID & FID & KID & FID & KID \\
        \midrule
        CLEVR-3           & 87.8 & 7.4 & 49.6 & 4.5 & 138.1 & 13.7 \\
        CLEVR-5-vbg        & 77.2 & 4.5 & 83.1 & 7.6 & 221.9 & 22.4 \\
        multi-dSprites   & 38.1 & 2.1 & 66.7 & 5.7 & 89.6 & 6.9 \\
        overlapping-polys & 49.5 & 3.7 & 75.9 & 7.5 & 81.1 & 7.9 \\
        \bottomrule
    \end{tabular}
\end{table}

\begin{table}
    \centering
    \caption{Ablation results showing performance on CLEVR-3 and CLEVR-5-vbg generation without the scene-level hyperprior parametrized by $\mathbf{y}$; compare with results from the full model in \tab{gen-results}. For full scene generations, the ablated model performs significantly worse than the complete model.}
    \label{tab:hyper-ablation}
    \vspace{4pt}
    \begin{tabular}{@{}l cc cc@{}}
        \toprule
         & \multicolumn{2}{c}{\textit{per-object}} & \multicolumn{2}{c}{\textit{full-scene}} \\
         & FID & KID & FID & KID \\
        \midrule
        CLEVR-3          & 98.3 & 8.3 & 66.1 & 5.8 \\
        CLEVR-5-vbg        & 66.6 & 3.9 & 92.9 & 8.6 \\
        \bottomrule
    \end{tabular}
\end{table}

\subsection{Image Generation}
We first demonstrate the ability of our model
to generate new high-quality images.

\paragraph{Qualitative results.}
\fig{gen-examples} shows illustrative examples of
new images being generated after training on the three CLEVR dataset variants,
respectively. More examples and results for other datasets can be found
in the supplemental material. 
One can see that our model can generate realistic-looking images of (up to the synthetic nature of the datasets) multi-object scenes with a variety of object appearances (size, colour, reflectivity), locations and depths.

When the training images for a generative model contain occlusions, one could expect the model to generate truncated objects, that only look correct when composed with other, specific objects occluding or adjoining them.
Our model does not exhibit this failure mode; the generated objects are complete and
look natural.
We attribute this to the combination of a low-dimensional appearance latent space
with the explicit image composition model.
The former discourages learning overly complex shapes, such as arbitrarily truncated objects. The latter ensures that such a low-dimensional representation suffices, as the object appearance decoder does not have to learn to to reconstruct objects at every possible location, but only at a single canonical position (centered in the canvas), with the compositor $G$ handling placement explicitly.

Also noteworthy is that our method generally produces objects with crisp, well-defined outlines---contrasting with typical behavior of VAEs.
We again attribute this to having an explicit representation of position; this means the appearance decoder does not have any incentive to average over different possible positions and therefore produce a blurry reconstruction.

\paragraph{Quantitative results.}
We report a quantitative evaluation of our model's ability to generate plausible images, based on the standard metrics Fr\'{e}chet inception distance (FID)~\citep{heusel17nips} and kernel inception distance (KID)~\citep{binkowski18iclr}.
These pass a large set of generated images through the CNN of \citet{szegedy16cvpr}, recording hidden layer feature activations;
they repeat this with a held-out set of ground-truth images, and finally measure how similar their feature activations are to those of the generated images.
We report KID~$\times 100$ instead of the raw values, for more convenient comparison.
We also give results for a baseline method that does not reason over objects---specifically, a deep convolutional VAE.
For fair evaluation, this has a similar architecture to our proposed model, but wider and deeper to generate full-size ($N \times N$) images, and with a latent capacity equal to the total of our object and background latent spaces.
To demonstrate that the individual objects generated by our model are indeed plausible in isolation, we evaluate the FID and KID between generated and ground-truth images containing only \textit{single} objects, but having trained our model on multi-object scenes (\emph{per-object} column).

The results in \tab{gen-results} show that our method significantly outperforms the VAE baseline, in spite of the latter having similar capacity (compare \textit{full-scene} and \textit{VAE baseline} columns in \tab{gen-results}).
This is true with respect to both FID and KID, and across all datasets.
Unfortunately, we cannot compare these metrics with existing works on object-level representations, as no published method demonstrates the generation of full scenes, as opposed to decomposition.


%
%

In \tab{hyper-ablation}, we evaluate the contribution to image quality of the scene-level prior parametrized by $\mathbf{y}$.
Specifically, we change the model to draw $\mathbf{z}_j$ and $d_j$ independently from standard Gaussian distributions, and $\theta_j$ from uniform Categorical distributions.
As expected, we see that the full-scene quality metrics worsen in the ablated case.
The model no longer learns co-occurrence relations between objects---for example, it cannot learn that intersections should be avoided.
Qualitative results in this ablated setting are given in the supplementary material.
%
The ablation has an inconsistent effect on the single-object quality metrics; for CLEVR-3, performance is lower, perhaps due to the ablated model being unable to capture dependence between position and appearance (e.g. due to perspective).
Conversely, for CLEVR-5-vbg, performance is higher, perhaps due to the scene-level prior encouraging the appearance model to use only a part of its latent space that yields coherent scenes without introducing difficult-to-model dependencies among objects.


\newcommand{\reconrowclevrthree}[2]{ 
        \includegraphics[scale=0.3]{recon-images/reconstruction-#1/#2_original.png} &
        \includegraphics[scale=0.3]{recon-images/reconstruction-#1/#2_final.png} &
        \includegraphics[scale=0.3]{recon-images/reconstruction-#1/#2_segmentation.png} &
        \includegraphics[scale=0.5]{recon-images/reconstruction-#1/#2_obj_0.png} &
        \includegraphics[scale=0.5]{recon-images/reconstruction-#1/#2_obj_1.png} &
        \includegraphics[scale=0.5]{recon-images/reconstruction-#1/#2_obj_2.png} &
        \includegraphics[scale=0.5]{recon-images/reconstruction-#1/#2_obj_3.png} &
        \includegraphics[scale=0.5]{recon-images/reconstruction-#1/#2_obj_4.png} &
        \includegraphics[scale=0.3]{recon-images/reconstruction-#1/#2_bg.png} &
        \includegraphics[scale=0.3]{recon-images/reconstruction-#1/#2_locations.png}  \\
}

\newcommand{\reconrowclevrfive}[2]{ 
	\includegraphics[scale=0.3]{recon-images/reconstruction-#1/#2_original.png} &
	\includegraphics[scale=0.3]{recon-images/reconstruction-#1/#2_final.png} &
	\includegraphics[scale=0.3]{recon-images/reconstruction-#1/#2_segmentation.png} &
	\includegraphics[scale=0.6]{recon-images/reconstruction-#1/#2_obj_0.png} &
	\includegraphics[scale=0.6]{recon-images/reconstruction-#1/#2_obj_1.png} &
	\includegraphics[scale=0.6]{recon-images/reconstruction-#1/#2_obj_2.png} &
	\includegraphics[scale=0.6]{recon-images/reconstruction-#1/#2_obj_3.png} &
	\includegraphics[scale=0.6]{recon-images/reconstruction-#1/#2_obj_4.png} &
	\includegraphics[scale=0.3]{recon-images/reconstruction-#1/#2_bg.png} &
	\includegraphics[scale=0.3]{recon-images/reconstruction-#1/#2_locations.png}  \\
}

\newcommand{\reconrowdsprites}[2]{ 
        \includegraphics[scale=0.45]{recon-images/reconstruction-#1/#2_original.png} &
		\includegraphics[scale=0.45]{recon-images/reconstruction-#1/#2_final.png} &
		\includegraphics[scale=0.45]{recon-images/reconstruction-#1/#2_segmentation.png} &
		\includegraphics[scale=0.75]{recon-images/reconstruction-#1/#2_obj_0.png} &
		\includegraphics[scale=0.75]{recon-images/reconstruction-#1/#2_obj_1.png} &
		\includegraphics[scale=0.75]{recon-images/reconstruction-#1/#2_obj_2.png} &
		\includegraphics[scale=0.75]{recon-images/reconstruction-#1/#2_obj_3.png} &
		\includegraphics[scale=0.75]{recon-images/reconstruction-#1/#2_obj_4.png} &
		\includegraphics[scale=0.45]{recon-images/reconstruction-#1/#2_bg.png} &
		\includegraphics[scale=0.45]{recon-images/reconstruction-#1/#2_locations.png}  \\
}

\newcommand{\reconrowpolygons}[2]{ 
	\includegraphics[scale=0.45]{recon-images/reconstruction-#1/#2_original.png} &
	\includegraphics[scale=0.45]{recon-images/reconstruction-#1/#2_final.png} &
	\includegraphics[scale=0.45]{recon-images/reconstruction-#1/#2_segmentation.png} &
	\includegraphics[scale=0.75]{recon-images/reconstruction-#1/#2_obj_0.png} &
	\includegraphics[scale=0.75]{recon-images/reconstruction-#1/#2_obj_1.png} &
	\includegraphics[scale=0.75]{recon-images/reconstruction-#1/empty.png} &
	\includegraphics[scale=0.75]{recon-images/reconstruction-#1/empty.png} &
	\includegraphics[scale=0.75]{recon-images/reconstruction-#1/empty.png} &
	\includegraphics[scale=0.45]{recon-images/reconstruction-#1/#2_bg.png} &
	\includegraphics[scale=0.45]{recon-images/reconstruction-#1/#2_locations.png}  \\
}

\newcommand{\reconrowsmall}[3]{ 
        \includegraphics[scale=0.4]{recon-images/reconstruction-#1/#2_original.png} &
        \includegraphics[scale=0.4]{recon-images/reconstruction-#1/#2_final.png} &
        \includegraphics[scale=0.4]{recon-images/reconstruction-#1/#2_segmentation.png} &
        \includegraphics[scale=0.7]{recon-images/reconstruction-#1/#2_obj_0.png} &
        \includegraphics[scale=0.7]{recon-images/reconstruction-#1/#2_obj_1.png} &
        \includegraphics[scale=0.4]{recon-images/reconstruction-#1/#2_locations.png} &
        \hspace{1pt} &
        \includegraphics[scale=0.4]{recon-images/reconstruction-#1/#3_original.png} &
        \includegraphics[scale=0.4]{recon-images/reconstruction-#1/#3_final.png} &
        \includegraphics[scale=0.4]{recon-images/reconstruction-#1/#3_segmentation.png} &
        \includegraphics[scale=0.7]{recon-images/reconstruction-#1/#3_obj_0.png} &
        \includegraphics[scale=0.7]{recon-images/reconstruction-#1/#3_obj_1.png} &
        \includegraphics[scale=0.4]{recon-images/reconstruction-#1/#3_locations.png} \\
}

\begin{figure*}[t]
    \centering
    \begin{tabular}{@{}cccccccccc@{}}
        original image & reconstruction & \parbox{5em}{predicted \\ segmentation} & \multicolumn{5}{c}{
            $\overbrace{\text{~furthest}\hspace{3cm}\text{nearest~}}^{\mbox{\normalsize\text{reconstructed objects}}}$
        } & background & locations \\
        \reconrowclevrthree{0}{107}
        \reconrowclevrthree{0}{371}
        \reconrowclevrfive{1}{041}
        \reconrowclevrfive{1}{081}
        \reconrowclevrfive{7}{983}
        \reconrowclevrfive{7}{511}
        \reconrowdsprites{5}{107}
        \reconrowdsprites{5}{200}
        \reconrowpolygons{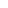}{043}
        \reconrowpolygons{6}{032}
    \end{tabular}
     \caption{Examples of scene decomposition by our model, on CLEVR-3 (top two rows), CLEVR-5, CLEVR-5-vbg, multi-dSprites and overlapping-polygons (bottom two rows).
     Note that relative depths and segmentation masks are inferred well, and objects are 
     recovered in full extent, even in the presence of strong occlusions.}  
    \label{fig:decomp-examples}
\end{figure*}

\subsection{Scene Decomposition}
Our model is trained with an encoder network, to reconstruct images; the encoder should correctly infer the decomposition of images into objects.
In this section, we evaluate the performance of the encoder to correctly segment images into individual objects and background, encode their appearances, depth and location, as well as the ability of the decoder to successfully reconstruct the scene. 

\begin{table}
    \centering
    \caption{Quantitative decomposition results applying our encoder network to held-out images. We report the IOU between ground-truth and predicted masks (both modal, mIOU, and amodal, aIOU), and the fraction of overlapping object pairs for which their depth ordering is correctly predicted (DPA). Higher is better for all metrics.}
    \label{tab:decomp-results}
    \vspace{4pt}
    \begin{tabular}{@{}l ccccc@{}}
        \toprule
         & mIOU & aIOU & DPA \\
        \midrule
        CLEVR-3           & 0.84 & 0.85 & 0.92 \\
        CLEVR-5            & 0.78 & 0.79 & 0.74 \\
        CLEVR-5-vbg         & 0.64 & 0.65 & 0.70 \\
        overlapping-polygons & 0.83 & 0.90 & 0.87 \\
        \bottomrule
    \end{tabular}
\end{table}

\paragraph{Qualitative results.}
\fig{decomp-examples} shows example images; more qualitative results for all datasets are in the supplementary material.
We see that our model reliably segments images into their constituent objects.
follow; this accords with the quantitative results in \tab{decomp-results}.
Even in cases with a large overlap between objects, it makes a reasonable prediction for the occluded pixels; this is particularly striking for overlapping-polygons, where the results are still qualitatively good for objects that are very highly occluded.
A numeric quantification of this effect is provided later in this section. 

While our model does not have explicit variables for object presence, we note that instead it simply sets the alpha channel to indicate the absence of an object or places the object outside of the visible area.

A noteworthy observation is that even in cases when the
results are not perfect, the model fails gracefully.
For example, in the fourth row of \fig{decomp-examples},
the brown object's shape is the back not recovered well,
but this does not negatively affect the object's position
or depths, and also the other objects are in fact
reconstructed well.

\paragraph{Quantitative results.}
Because the datasets are generated synthetically, we are able
to evaluate several metrics that quantify the ability of our
model to correctly parse images of scenes into their objects.
For each object $j$, our model estimates a depth $d_j$ and an amodal alpha mask placed into the full image, $\alpha^*_j$ (\eqn{alpha-placement}).
From these, we construct
(i) a modal segmentation mask of the image, by labelling each pixel with the nearest object having $\alpha^*_j > 0.3$ (or `background' if there is no such object), and 
(ii) an amodal segmentation mask for each object, consisting of all pixels with $\alpha^*_j > 0.3$.
Our metrics are then:
\begin{enumerate}
    \item modal segmentation intersection-over-union (\textbf{mIOU}): we match each ground-truth object to the most-overlapping reconstructed object (in terms of intersection-over-union---IOU---of the modal masks) using the Hungarian algorithm~\citep{kuhn55nlq}. Then, for each matched pair of ground-truth and reconstruction, we measure the IOU between their respective masks.
    \item amodal segmentation intersection-over-union (\textbf{aIOU}): this is identical to mIOU, except the amodal masks are used for the ground-truth and reconstruction. Thus, this metric measures how accurately the extents of both visible and hidden parts of objects are predicted.
    \item pairwise accuracy of depth-ordering (\textbf{DPA}): for each pair of objects with at least 30 pixels overlapping, we check whether their depth ordering predicted by our model matches the ground-truth ordering; we report the fraction of pairs with correct predictions.
\end{enumerate}


%
The results confirm our observation from the previous section that
the model is able to correctly segment the hidden parts of object,
in spite of not receiving any segmentation annotations during training:
the mean amodal IOU is 0.90 for overlapping-polygons and 0.85 CLEVR-3.
This is particularly noteworthy given that no prior works support amodal segmentation in the fully-unsupervised setting.
%
The worst-performing dataset is CLEVR-5-vbg, which has a large number of small objects, and a varying background; however, even in this case, the modal and amodal IOUs are around 0.65.

The model predicts depth ordering correctly in 92\% of cases on CLEVR-3, and 87\% on overlapping-polygons; again the result is somewhat lower on CLEVR-5-vbg, at 70\% (compared to chance at 50\%).
Note that when two objects do not overlap, there is no motivation for any model to predict the correct ordering---either results in the same image.
In the supplementary material, we examine how the depth accuracy varies with the degree of overlap between objects.

\subsection{Disentangled Interpolations}

\begin{figure}
    \centering
    \includegraphics[width=0.18\linewidth]{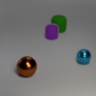}
    \includegraphics[width=0.18\linewidth]{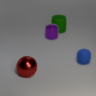}
    \includegraphics[width=0.18\linewidth]{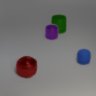}
    \includegraphics[width=0.18\linewidth]{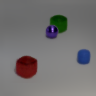}
    \includegraphics[width=0.18\linewidth]{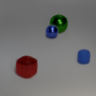}
    \\
    \includegraphics[width=0.18\linewidth]{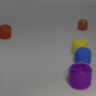}
    \includegraphics[width=0.18\linewidth]{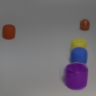}
    \includegraphics[width=0.18\linewidth]{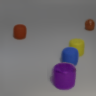}
    \includegraphics[width=0.18\linewidth]{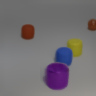}
    \includegraphics[width=0.18\linewidth]{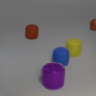}
     \\
    \includegraphics[width=0.18\linewidth]{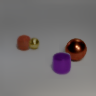}
    \includegraphics[width=0.18\linewidth]{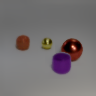}
    \includegraphics[width=0.18\linewidth]{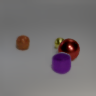}
    \includegraphics[width=0.18\linewidth]{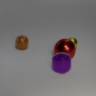}
    \includegraphics[width=0.18\linewidth]{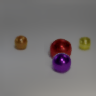}
    \caption{Disentangled interpolations in latent space. For each row, we interpolate between two images in latent space.
    \textit{Top row}: we keep the object positions fixed as their appearances vary/
    \textit{Middle row}: we vary the object positions, while keeping $\mathbf{y}$ fixed/
    \textit{Bottom row}: we allow all parameters to vary jointly.
    We see that the structured latent space allows disentangling of position and appearance variations.}
    \label{fig:interpolations}
\end{figure}

We now show that our structured latent space allows varying different factors in a disentangled fashion.
Specifically, we encode two images, and then interpolate linearly between them in latent space.
Due to the factoring of the latent space, we may choose whether to interpolate only object positions or appearances, or both; we show one example of each case.
In the top row of \fig{interpolations}, we fix the elements of $\mathbf{y}$ that influence
the object positions $\theta_j$, varying only those that affect
the object appearance embeddings.
Consequently, we see that the individual object appearances change, but they remain perfectly stationary, and the background is unaffected.
In the middle row, we do the opposite---fixing the appearances, but varying the positions.
Our model explicitly allows the positions to influence appearance (to account for perspective effects), yet explicit manipulation of the latent space
does not lead to visual artefacts, indicating that the learned
representation is sufficiently disentangled. 
In the bottom row, we allow both positions and appearances to vary, meaning the scene changes in both composition and layout, yet still it maintains a plausible appearance throughout.

\section{Conclusion}
\label{sec:conclusion}

We have presented a new generative model of images, that explicitly models their composition from objects with positions, depths, and amodal segmentation masks.
This model allows sampling images showing both isolated objects and coherent scenes.
Moreover, we have shown that its encoder network can segment images, including amodal segmentation and pixel prediction for occluded object parts.
It explicitly reasons over the occlusion ordering of objects, and is able to successfully predict this.
As future work, it would be interesting to combine our method with more powerful, iterative inference techniques, such as that used by \citet{greff19icml}.
%
Finally, we hope that the combination of deep, black-box neural networks with classical, human-specified directed probabilistic models, is fruitful in other domains, where data is too complex to model explicitly, yet has some predictable structure that can be captured in a prior and exploited to ease learning.

\bibliography{shortstrings,calvin,vggroup}
\bibliographystyle{icml2020}

\appendix
\onecolumn

\begin{table}
	\centering
	\caption{Reconstruction mean squared error and its sample standard deviation ($N=5$)}
	\label{tab:stds}
	\vspace{4pt}
    \begin{tabular}{@{}lcc@{}}
		\toprule
		& MSE ($\times 10^{-4}$) & STD ($\times 10^{-4}$) \\
		\midrule
		CLEVR-3           & 6.0 & 0.53 \\
		CLEVR-5            & 8.1 & 0.09 \\
		CLEVR-5-vbg        & 7.4 & 0.57 \\
		\bottomrule
	\end{tabular}
\end{table}

\begin{table*}[t]
	\begin{center}
		\caption{
            Encoder and decoder network architectures used for each dataset.
            $F_\mathrm{obj}$, $F_\mathrm{bg}$, and $\xi$ are the per-object, background and scene-level decoders defined in Section~3 of the paper.
            $\text{enc}_\text{scene}$ is the encoder parametrizing $Q(\mathbf{y} \given \mathbf{z}_{1 \ldots J} ,\, \theta_{1 \ldots J} ,\, d_{1 \ldots J})$; 
            $\text{enc}_\text{obj}$ is the encoder parametrizing $Q(\mathbf{z}_j ,\, d_j \given \theta_j ,\, \mathbf{x})$; 
            $\text{enc}_\text{pos}$ is the encoder parametrizing $Q(\theta_j \given \mathbf{x})$; 
            $\text{enc}_\text{bg}$ is the encoder parametrizing $Q(\mathbf{z}_\mathrm{bg} \given \mathbf{x})$.
            We denote convolutional layers with `c', transpose convolutional with `c\textsuperscript{T}', and fully-connected layers with `d'.
            The number after denotes the number of channels or units (if `r' it is a residual layer with the same number of channels as the previous) and the last letter denotes the activation function (`r' for ReLU and `e' for ELU).
            `k` and `s` denote convolution kernel size and stride respectively; if omitted kernel size is assumed to be 3 and stride 2.
            Bilinear upsampling is denoted by `up' followed by the factor.
            Max-pooling ($2 \times 2$) is denoted by `mp'.
            Batch normalization is denoted by `b' and group normalization by `g'.
            \emph{Source code and trained models will also be released soon.}
        }
        \label{tab:our-arch}
        \vspace{6pt}
        \begin{tabular}{ | l | c| }
            \hline
            & \textbf{CLEVR-3} \\
            \hline
            $\text{enc}_\text{scene}$ & d128e-dre-dre-d32e \\
            $\text{enc}_\text{obj}$ & c16e-b-c32e-b-c64e-b-c128e-b-c256e-b-c512e-c1024e-b-c2048e-b-d66 \\
            $\text{enc}_\text{bg}$ & c16r-b-c32r-b-c48r-b-c64r-b-c96r-f-d128r-b-d6 \\
            $\text{enc}_\text{pos}$ & c48r-g-mp-c64r-g-mp-c96r-g-mp-c128r-g-mp-c128r-up2-c128r-g-up2-c96r-g-up2-c64r-g-up2-c48r-c3k1 \\
            $\xi$ & d128e-dre-dre-d42e \\
            $F_\mathrm{obj}$ & d128e-d3000e-c\textsuperscript{T}128e-c\textsuperscript{T}64e-c\textsuperscript{T}32k5s1e-c\textsuperscript{T}4k1s1s \\
            $F_\mathrm{bg}$ & d144e-up2-c\textsuperscript{T}3e-up4-c\textsuperscript{T}3e-up4-c\textsuperscript{T}3k1s1s \\
            \hline
            \hline
            & \textbf{CLEVR-5} and \textbf{CLEVR-5-vbg} \\
            \hline
            $\text{enc}_\text{scene}$ & d128e-dre-dre-d32e \\
            $\text{enc}_\text{obj}$ & c16e-b-c32e-b-c64e-b-c128e-b-c256e-b-c512e-c1024e-b-c2048e-b-d66 \\
            $\text{enc}_\text{bg}$ & c16r-b-c32r-b-c48r-b-c64r-b-c96r-f-d128r-b-d6 \\
            $\text{enc}_\text{pos}$ & c48r-g-mp-c64r-g-mp-c96r-g-mp-c128r-g-mp-c128r-up2-c128r-g-up2-c96r-g-up2-c64r-g-up2-c48r-c3k1 \\
            $\xi$ & d128e-dre-dre-d68e \\
            $F_\mathrm{obj}$ & d128e-d1920e-c\textsuperscript{T}128e-c\textsuperscript{T}64e-c\textsuperscript{T}32k5s1e-c\textsuperscript{T}4k1s1s \\
            $F_\mathrm{bg}$ & d144e-up2-c\textsuperscript{T}3e-up4-c\textsuperscript{T}3e-up4-c\textsuperscript{T}3k1s1s \\
            \hline
            \hline
            & \textbf{overlapping-polygons} \\
            \hline
            $\text{enc}_\text{scene}$ & d128e-dre-dre-d32e \\
            $\text{enc}_\text{obj}$ & c16e-b-c32-e-b-c64-e-b-c128-e-b-c256-e-b-c512-e-c1024-e-b-c2048-e-b-d66 \\
            $\text{enc}_\text{bg}$ & c16r-b-c32r-b-c48r-b-c64r-b-c96r-f-d128r-b-d4 \\
            $\text{enc}_\text{pos}$ & c48r-g-mp-c64r-g-mp-c96r-g-mp-c128r-g-mp-c128r-up2-c128r-g-up2-c96r-g-up2-c64r-g-up2-c48r-c3k1 \\
            $\xi$ & d128e-dre-dre-d28e \\
            $F_\mathrm{obj}$ & d128e-d1080e-c\textsuperscript{T}128e-c\textsuperscript{T}64e-c\textsuperscript{T}32k5s1e-c\textsuperscript{T}4k1s1s \\
            $F_\mathrm{bg}$ & d64e-up2-c\textsuperscript{T}3e-up4-c\textsuperscript{T}3e-up4-c\textsuperscript{T}3k1s1s \\
            \hline
            \hline
            & \textbf{multi-dSprites} \\
            \hline
            $\text{enc}_\text{scene}$ & d128e-dre-dre-d32e \\
            $\text{enc}_\text{obj}$ & c32e-c64k1s1e-crk1s1e-g-c64e-c128k1s1e-crk1s1e-g-c256e-c256k1s1e-crk1s1e-g-d14 \\
            $\text{enc}_\text{bg}$ & c16r-b-c32r-b-c48r-b-c64r-b-c96r-f-d128r-b-d4 \\
            $\text{enc}_\text{pos}$ & c48r-g-mp-c64r-g-mp-c96r-g-mp-c128r-g-mp-c128r-up2-c128r-g-up2-c96r-g-up2-c64r-g-up2-c48r-c3k1 \\
            $\xi$ & d128e-dre-dre-d54e \\
            $F_\mathrm{obj}$ & d2048e-c128s1e-crs1e-up2-c96s1e-crs1e-crs1e-up2-c48s1e-crs1e-crs1e-up2-crs1e-crs1e-c4s1 \\
            $F_\mathrm{bg}$ & d64e-up2-c\textsuperscript{T}3e-up4-c\textsuperscript{T}3e-up4-c\textsuperscript{T}3k1s1s \\
            \hline
        \end{tabular}
    \end{center}
\end{table*}

\begin{table*}
    \centering
    \caption{Network architectures for the baseline VAE encoder and decoder. For notation, see \tab{our-arch}. The final encoder layer has channel count equal to the total object-level latent dimensionality of our model for the corresponding dataset, i.e. $D_\mathrm{obj} \times J + D_\mathrm{bg}$.
    \emph{Source code and trained models will also be released soon.}}
    \label{tab:baseline-arch}
    \vspace{6pt}
    \begin{tabular}{|l|c|}
				\hline
                & \textbf{CLEVR-3}, \textbf{CLEVR-5}, and \textbf{CLEVR-5-vbg} \\
				\hline
				\textbf{encoder} & c32e-c64k1s1e-crk1s1e-g-c64e-c128k1s1e-crk1s1e-g-c256e-c256k1s1e-crk1s1e-g-c256e-g-d \\
				\textbf{decoder} & d4608e-c128s1e-crs1e-up2-c96s1e-crs1e-crs1e-up2-c48s1e-crs1e-crs1e-up2-crs1e-crs1e-c4k1s1 \\
				\hline
                \hline
                & \textbf{multi-dSprites} and \textbf{overlapping-polygons} \\
				\hline
				\textbf{encoder} & c32e-c64k1s1e-crk1s1e-g-c64e-c128k1s1e-crk1s1e-g-c256e-c256k1s1e-crk1s1e-g-c256e-g-d \\
				\textbf{decoder} & d2048e-c128s1e-crs1e-up2-c96s1e-crs1e-crs1e-up2-c48s1e-crs1e-crs1e-up2-crs1e-crs1e-c4k1s1 \\
				\hline
    \end{tabular}
\end{table*}

\section{Network Architectures}
\label{sec:architectures}

\tab{our-arch} specifies the network architectures for each component of our model.
Note that these differ slightly between datasets to account for their different image sizes $N$, and different numbers of object slots $J$.
Specifically, we use $J$=$5$ for \textit{CLEVR-5} and \textit{CLEVR-5-vbg}, $J$=$3$ for \textit{CLEVR-3}, $J$=$4$ for \textit{multi-dSprites}, and $J$=$2$ for \textit{overlapping-polygons}---corresponding in each case to the maximum number of objects present. For \textit{multi-dSprites}, we observed that model sometimes finds it difficult to reproduce sharp shapes, such as hearts---a problem also encountered by \citet{greff19icml}. We resolved this problem by (i) adding residual connections in the encoder and decoder, and (ii) incorporating attentive encoding, so $\text{enc}_\text{obj}$ is run once per object with an $M \times M$ weighted average crop as input, as described in Section~4 of the main paper.

\tab{baseline-arch} specifies the network architectures used for the baseline VAE experiments.

\section{Training and Hyperparameters}

Training was performed for $500$ epochs with a batch size of $100$ with the last 10 epochs used to train $\text{enc}_\text{scene}$ and $\xi$. The dataset consisted of $10^{5}$ samples, $9 \times 10^{4}$ used for training leaving remained for evaluation. We used Adam~\citep{kingma15iclr} for optimization, with a constant learning rate of $5 \times 10^{-4}$. Images were normalized to values between 0 and 1 and their likelihood function was a Normal distribution with the mean predicted by the model and standard deviation of $0.01 / \sqrt 3$ for epochs 1--150, and $0.01 / \sqrt 5$ thereafter.
A weight in front of the KL-divergence for the hyperprior $\mathbf{y}$ was $0.01$. At epoch $i$, the temperature of the Gumbel-Softmax distribution on $\theta_j$ was $0.3 \times 2^{0.0001 \times i}$.

Note that training was ended at 500 epochs due to time constraints.
The models had not fully converged at this point---the loss, and especially the visual image quality, continue to improve significantly until at least 2000 epochs.
However, the decomposition into objects and inference of their positions converges within the first few epochs, with slow improvement to the reconstructed pixels thereafter.

\section{Datasets}

\subsection{CLEVR}

We generated all CLEVR datasets using a modified version of the code from \cite{johnson17cvpr}. Each contains $10^{5}$ images of size (96, 96, 3). Scenes in \textit{CLEVR-5} and \textit{CLEVR-5-vbg} contain 3, 4 or 5 objects; scenes in \textit{CLEVR-3} contain 2 or 3 objects. Each object can be a cube, cylinder or sphere, with one of two sizes, eight colours, and two materials, as well as a position and azimuth. For \textit{CLEVR-3} we ensure a higher degree of overlap (more challenging to learn) by increasing the object sizes and allowing images with fewer visible pixels per object than for \textit{CLEVR-5} (10 vs. 50). For \textit{CLEVR-5-vbg}, we randomly sampled the diffuse reflectivity of the gray background from one of six logarithmically-spaced values.

\subsection{Multi-dSprites}

The original \textit{multi-dSprites} dataset of \citet{burgess19arxiv} contains $10^6$ images of shape (64, 64, 3), each showing 2, 3 or 4 shapes such as hearts and squares, with randomly-sampled positions, rotations, and colors, overlaid from back to front in a random order.
We use their original images, but take only the first $10^5$ of these, discarding the remainder.
To evaluate the single-object FID/KID metrics, we generate pseudo ground-truth images by selecting regions that are of uniform color, and do not touch any other non-black pixels (so their extent is unambiguous).
Note that this creates a slight bias in evaluation compared with regenerating single-object images from their original distribution.

\subsection{Overlapping-polygons}

\textit{overlapping-polygons} contains $10^{5}$ images of size (64, 64, 3), each showing 2 filled polygons.
Each polygon has 3--6 (inclusive) edges, a radius uniformly sampled between 7.5 and 12.5 pixels, and a color generated by sampling RGB values uniformly from [0, 1].
Its center is offset from the center of the image by up to 5 pixels in each direction (again uniformly sampled); this results in overlapping objects in nearly all images.
The polygons are drawn from back to front over a black background.

\section{Additional Results}

Additional results for decomposition on overlapping-polygons are given in \fig{decomp-examples-overlapping}.
Note that the model is reliably successful in amodally predicting the full extent of occluded objects; moreover, the relative object depths are always inferred correctly.
Additional results for generation and decomposition on all CLEVR variants and multi-dsprites are shown in Figures \ref{fig:decomp-examples-dsprites}--\ref{fig:gen-examples-clevr3}.
Note in particular that for multi-dSprites, our model outputs sharp, accurate outlines, unlike the prior work of \citet{greff19icml} (compare their Figure~17 with our \fig{decomp-examples-dsprites}).

\tab{stds} shows the mean squared reconstruction error (MSE) on evaluation images for the three variants of CLEVR; this is an average over 5 runs with different random seeds, and we also give the standard deviations.
Note that the standard deviations are typically small, indicating that the model reliably converges to similar performance regardless of random initialization.

\subsection{Depth accuracy at differing overlap thresholds}

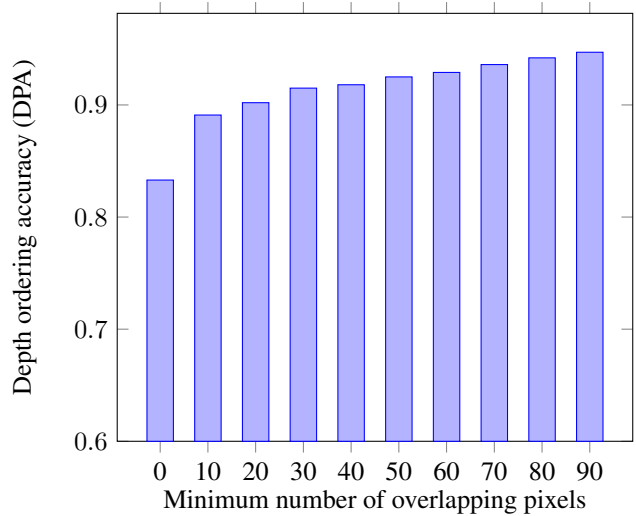
\begin{figure}\centering
\pgfplotstableread[row sep=\\,col sep=&]{
	pixels & accuracy  \\
	0       & 0.833   \\
	10      & 0.891  \\
	20      & 0.902   \\
	30      & 0.915   \\
	40      & 0.918   \\
	50      & 0.925   \\
	60      & 0.929   \\
	70      & 0.936   \\
	80      & 0.942   \\
	90      & 0.947   \\
}\mydata

\begin{tikzpicture}
\begin{axis}[
ybar,
symbolic x coords={0,5,10,15,20,25,30,35,40,45,50,55,60,65,70,75,80,85,90,95},
xtick=data,
ymin=0.6,
ylabel={Depth ordering accuracy (DPA)},
xlabel={Minimum number of overlapping pixels}
]
\addplot table[x=pixels,y=accuracy]{\mydata};
\end{axis}
\end{tikzpicture}
    \caption{Accuracy of predicting the pairwise depth ordering of object pairs (DPA, $y$-axis) 
    that overlap by at least by a certain threshold ($x$-axis)}
\label{fig:dpa}
\end{figure}

As described in the paper, we measure pairwise accuracy of depth-ordering \textbf{(DPA)} by taking pairs of objects which have at least 30 pixels overlapping and checking whether their depth ordering predicted by our model matches the ground-truth. 
In \fig{dpa}, we measure the influence of this threshold on the DPA score, for the CLEVR-3 dataset.
We observe that as the overlap threshold increases, the DPA score increases; thus, for objects that have a larger degree of overlap, our model infers their ordering more successfully, increasing to 94.7\% of pairs when the threshold is set to 90 pixels.
When there is no overlap between objects, the model does not receive any training signal to influence their predicted depths, as the depths do not affect the final appearance.
Consequently, we see that the depth prediction accuracy is markedly lower, at 83.3\%.

\subsection{Ablation of hyperprior}

\fig{gen-without-y} shows generations without the scene-level hyperprior parametrized by $\mathbf{y}$; we see this results in unrealistic visual scenes.
This is because without the hyperprior, the model assumes objects appear independently (as do the models of \citet{greff19icml,eslami16nips})---which is not true in reality.
In contrast, with the hyperior enabled, neither our generative nor inference model assume that objects are independent. This allows inference network to not explain the same object twice and, similarly to humans, understand the same object differently depending on the context in which it appears.
By incorporating the hyperprior in our model (as in \fig{gen-examples-clevr3}), the generative process captures the relationships between objects and the spatial organization that is imposed on the scene by physical world, such as non-intersection, placement on a common `ground' plane, and perspective effects on appearance.

\subsection{Speed and comparison to other models}

Spatial and depth invariant object placing allows us to reuse networks exploiting the fact that visual scenes are composed of objects. This is manifested in better training, faster inference and generation speed as well as a small memory consumption. Our attentive inference and generation has the fastest computational complexity among models that are known to us. On average, a forward pass through a model (inference and generation) of a (64, 64) image batch of size 100 takes 0.25s of CPU clock time on NVIDIA GeForce GTX1080Ti (that we used for training). This makes it possible to use the model for real-time applications and train it within a day.

\newcommand{\reconrowoverlapping}[2]{ 
	\includegraphics[scale=0.85]{recon-images/reconstruction-#1/#2_original.png} &
	\includegraphics[scale=0.85]{recon-images/reconstruction-#1/#2_final.png} &
	\includegraphics[scale=0.85]{recon-images/reconstruction-#1/#2_segmentation.png} &
	\includegraphics[scale=0.85]{recon-images/reconstruction-#1/#2_obj_0.png} &
	\includegraphics[scale=0.85]{recon-images/reconstruction-#1/#2_obj_1.png} &
	\includegraphics[scale=0.85]{recon-images/reconstruction-#1/#2_bg.png} &
	\includegraphics[scale=0.85]{recon-images/reconstruction-#1/#2_locations.png}  \\
}

\begin{figure*}[t]
	\centering
	\begin{tabular}{@{}cccccccccc@{}}
		original image & reconstruction & \parbox{5em}{predicted \\ segmentation} & \multicolumn{2}{c}{
			$\overbrace{\text{~furthest}\hspace{1cm}\text{nearest~}}^{\mbox{\normalsize\text{reconstructed objects}}}$
		} & background & locations \\
		\reconrowoverlapping{6}{000}
		\reconrowoverlapping{6}{001}
		\reconrowoverlapping{6}{002}
		\reconrowoverlapping{6}{003}
		\reconrowoverlapping{6}{004}
		\reconrowoverlapping{6}{005}
		\reconrowoverlapping{6}{006}
		\reconrowoverlapping{6}{007}
		\reconrowoverlapping{6}{008}
		\reconrowoverlapping{6}{009}
	\end{tabular}
	\caption{Reconstruction of the first ten (i.e. random) samples from overlapping-polygons evaluation dataset. Note how successful is the model's amodal perception of occluded objects. Also note that when only one object is seen in the original image, the model places a second object outside the visible area. Finally, note that all relative object depths are inferred correctly.}
	\label{fig:decomp-examples-overlapping}
\end{figure*}

\begin{figure*}
	\centering
	\includegraphics[width=0.18\linewidth]{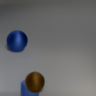}
	\includegraphics[width=0.18\linewidth]{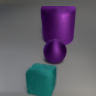}
	\includegraphics[width=0.18\linewidth]{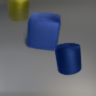}
	\includegraphics[width=0.18\linewidth]{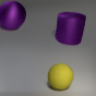}
	\includegraphics[width=0.18\linewidth]{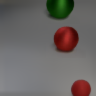}
	\includegraphics[width=0.18\linewidth]{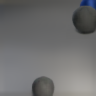}
	\includegraphics[width=0.18\linewidth]{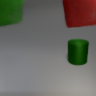}
	\includegraphics[width=0.18\linewidth]{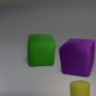}
	\includegraphics[width=0.18\linewidth]{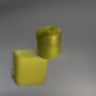}
	\includegraphics[width=0.18\linewidth]{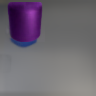}
	\includegraphics[width=0.18\linewidth]{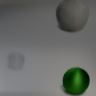}
	\includegraphics[width=0.18\linewidth]{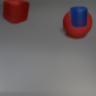}
	\includegraphics[width=0.18\linewidth]{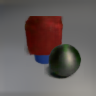}
	\includegraphics[width=0.18\linewidth]{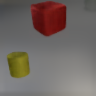}
	\includegraphics[width=0.18\linewidth]{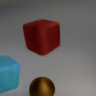}
	\includegraphics[width=0.18\linewidth]{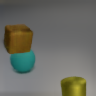}
	\includegraphics[width=0.18\linewidth]{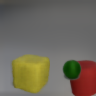}
	\includegraphics[width=0.18\linewidth]{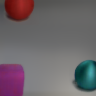}
	\includegraphics[width=0.18\linewidth]{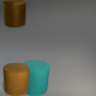}
	\includegraphics[width=0.18\linewidth]{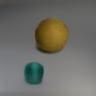}
	\includegraphics[width=0.18\linewidth]{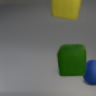}
	\includegraphics[width=0.18\linewidth]{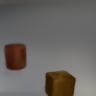}
	\includegraphics[width=0.18\linewidth]{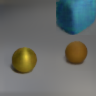}
	\includegraphics[width=0.18\linewidth]{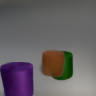}
	\includegraphics[width=0.18\linewidth]{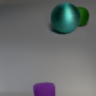}

	\caption{CLEVR-3 generations \textbf{without the scene-level hyperprior parameterized by $\mathbf{y}$}. This illustrates that it is necessary for a generative process to capture the relationships between objects and the spatial organization that is imposed on the scene by physical world, such as non-intersection, placement on a common ‘ground’ plane, and perspective effects on appearance.}
	\label{fig:gen-without-y}
\end{figure*}

\newcommand{\genrowoverlapping}[2]{ 
	\includegraphics[scale=0.9]{gen-images/#1/#2_final.png} &
	\includegraphics[scale=0.9]{gen-images/#1/#2_segmentation.png} &
	\includegraphics[scale=0.9]{gen-images/#1/#2_obj_0.png} &
	\includegraphics[scale=0.9]{gen-images/#1/#2_obj_1.png} &
	\includegraphics[scale=0.9]{gen-images/#1/#2_bg.png}  \\
}

\renewcommand{\reconrowdsprites}[2]{ 
	\includegraphics[scale=1.0]{recon-images/reconstruction-#1/#2_original.png} &
	\includegraphics[scale=1.0]{recon-images/reconstruction-#1/#2_final.png} &
	\includegraphics[scale=1.0]{recon-images/reconstruction-#1/#2_segmentation.png} &
	\includegraphics[scale=1.0]{recon-images/reconstruction-#1/#2_obj_0.png} &
	\includegraphics[scale=1.0]{recon-images/reconstruction-#1/#2_obj_1.png} &
	\includegraphics[scale=1.0]{recon-images/reconstruction-#1/#2_obj_2.png} &
	\includegraphics[scale=1.0]{recon-images/reconstruction-#1/#2_obj_3.png} &
	\includegraphics[scale=1.0]{recon-images/reconstruction-#1/#2_bg.png} &
	\includegraphics[scale=1.0]{recon-images/reconstruction-#1/#2_locations.png}  \\
}

\begin{figure*}[t]
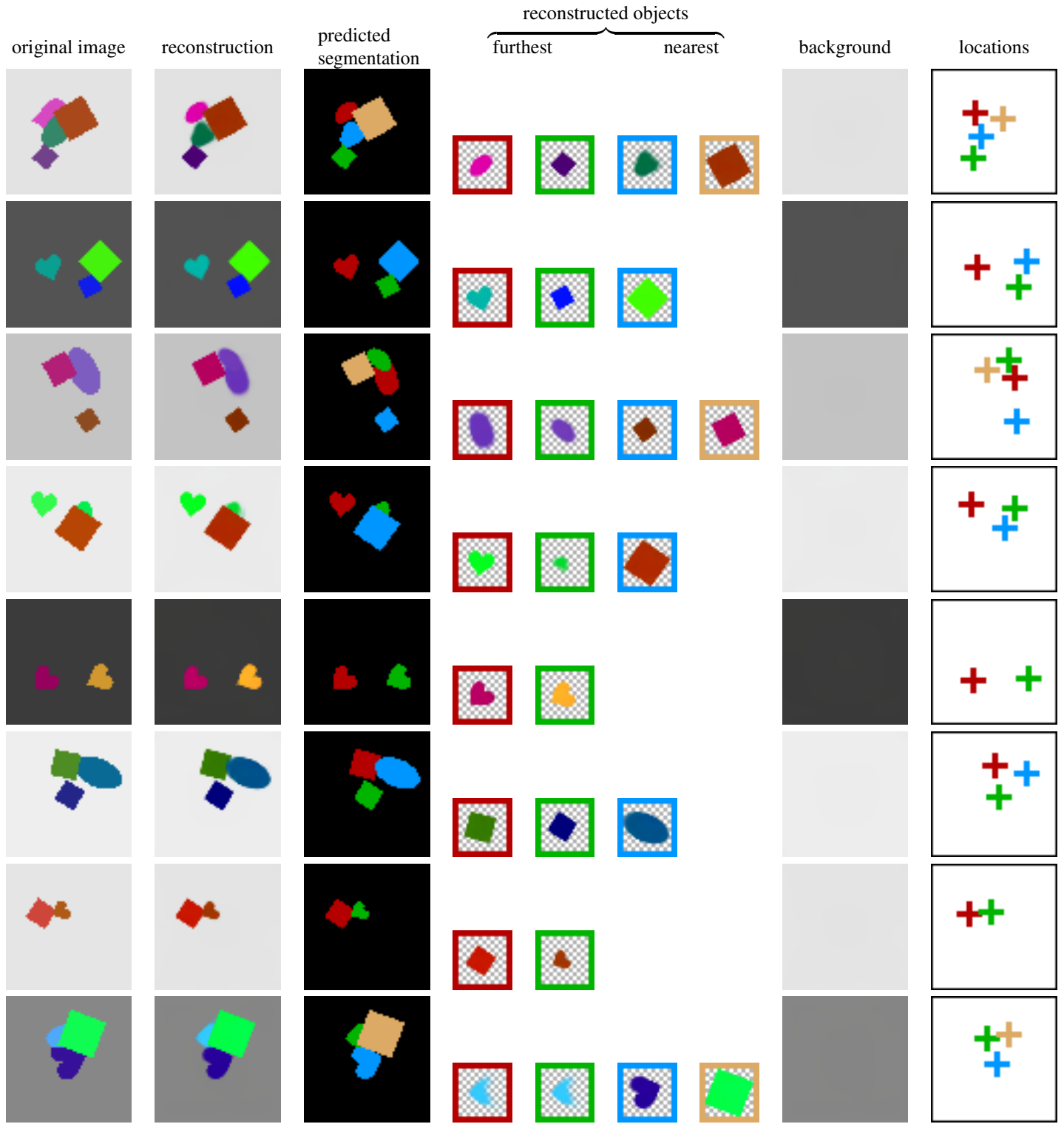

	\centering
	\begin{tabular}{@{}ccccccccc@{}}
		original image & reconstruction & \parbox{5em}{predicted \\ segmentation} & \multicolumn{4}{c}{
			$\overbrace{\text{~furthest}\hspace{2cm}\text{nearest~}}^{\mbox{\normalsize\text{reconstructed objects}}}$
		} & background & locations \\
		\reconrowdsprites{5}{000}
		\reconrowdsprites{5}{011}
		\reconrowdsprites{5}{002}
		\reconrowdsprites{5}{003}
		\reconrowdsprites{5}{004}
		\reconrowdsprites{5}{005}
		\reconrowdsprites{5}{006}
		\reconrowdsprites{5}{007}
	\end{tabular}
	\caption{multi-dSprites decompositions.}
	\label{fig:decomp-examples-dsprites}
\end{figure*}

\newcommand{\genrowdsprites}[2]{ 
	\includegraphics[scale=1.0]{gen-images/#1/#2_final.png} &
	\includegraphics[scale=1.0]{gen-images/#1/#2_segmentation.png} &
	\includegraphics[scale=1.0]{gen-images/#1/#2_obj_0.png} &
	\includegraphics[scale=1.0]{gen-images/#1/#2_obj_1.png} &
	\includegraphics[scale=1.0]{gen-images/#1/#2_obj_2.png} &
	\includegraphics[scale=1.0]{gen-images/#1/#2_obj_3.png} &
	\includegraphics[scale=1.0]{gen-images/#1/#2_bg.png} &
	\includegraphics[scale=1.0]{gen-images/#1/#2_locations.png} \\
}

\begin{figure*}[t!]
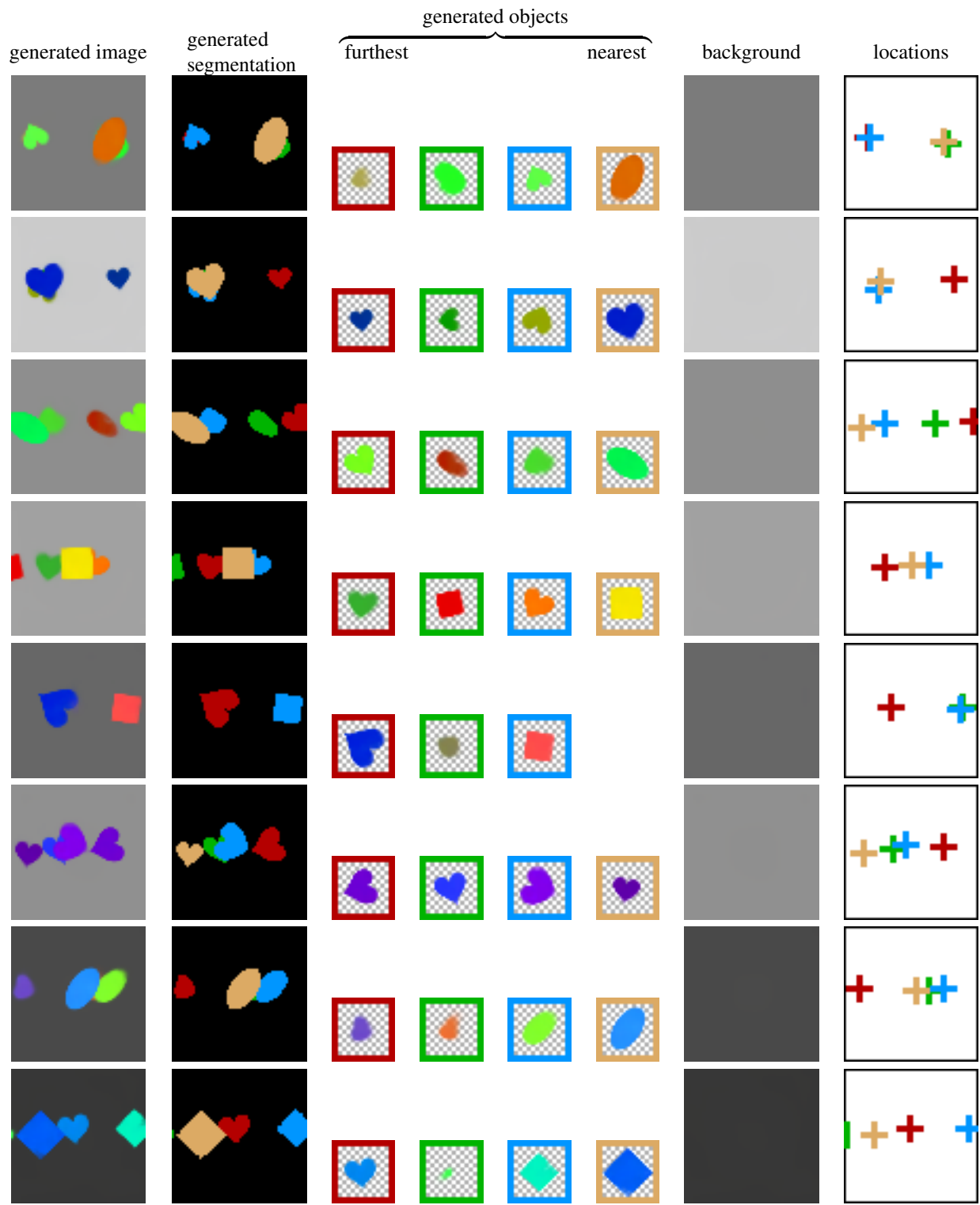

	\centering
	\begin{tabular}{@{}cccccccc@{}}
		generated image & \parbox{5em}{generated \\ segmentation} & \multicolumn{4}{c}{
			$\overbrace{\text{~furthest}\hspace{3cm}\text{nearest~}}^{\mbox{\normalsize\text{generated objects}}}$
		} & background & locations \\
		\genrowdsprites{5}{000}
		\genrowdsprites{5}{001}
		\genrowdsprites{5}{002}
		\genrowdsprites{5}{299}
		\genrowdsprites{5}{080}
		\genrowdsprites{5}{188}
		\genrowdsprites{5}{006}
		\genrowdsprites{5}{007}
	\end{tabular}
	\\
	\caption{multi-dSprites generations.}
	\label{fig:gen-examples-dsprites}
\end{figure*}

\renewcommand{\reconrowclevrfive}[2]{ 
	\includegraphics[scale=0.6]{recon-images/reconstruction-#1/#2_original.png} &
	\includegraphics[scale=0.6]{recon-images/reconstruction-#1/#2_final.png} &
	\includegraphics[scale=0.6]{recon-images/reconstruction-#1/#2_segmentation.png} &
	\includegraphics[scale=0.6]{recon-images/reconstruction-#1/#2_obj_0.png} &
	\includegraphics[scale=0.6]{recon-images/reconstruction-#1/#2_obj_1.png} &
	\includegraphics[scale=0.6]{recon-images/reconstruction-#1/#2_obj_2.png} &
	\includegraphics[scale=0.6]{recon-images/reconstruction-#1/#2_obj_3.png} &
	\includegraphics[scale=0.6]{recon-images/reconstruction-#1/#2_obj_4.png} &
	\includegraphics[scale=0.6]{recon-images/reconstruction-#1/#2_bg.png} &
	\includegraphics[scale=0.6]{recon-images/reconstruction-#1/#2_locations.png}  \\
}

\begin{figure*}[t]
	\centering
	\begin{tabular}{@{}cccccccccc@{}}
		original image & reconstruction & \parbox{5em}{predicted \\ segmentation} & \multicolumn{5}{c}{
			$\overbrace{\text{~furthest}\hspace{3cm}\text{nearest~}}^{\mbox{\normalsize\text{reconstructed objects}}}$
		} & background & locations \\
		\reconrowclevrfive{1}{056}
		\reconrowclevrfive{1}{011}
		\reconrowclevrfive{1}{074}
		\reconrowclevrfive{1}{089}
		\reconrowclevrfive{1}{150}
		\reconrowclevrfive{1}{005}
		\reconrowclevrfive{1}{006}
		\reconrowclevrfive{1}{150}
	\end{tabular}
	\caption{CLEVR-5 decompositions.}
	\label{fig:decomp-examples-clevr5}
\end{figure*}

\newcommand{\genrowclevrfive}[2]{ 
	\includegraphics[scale=0.6]{gen-images/#1/#2_final.png} &
	\includegraphics[scale=0.6]{gen-images/#1/#2_segmentation.png} &
	\includegraphics[scale=0.6]{gen-images/#1/#2_obj_0.png} &
	\includegraphics[scale=0.6]{gen-images/#1/#2_obj_1.png} &
	\includegraphics[scale=0.6]{gen-images/#1/#2_obj_2.png} &
	\includegraphics[scale=0.6]{gen-images/#1/#2_obj_3.png} &
	\includegraphics[scale=0.6]{gen-images/#1/#2_obj_4.png} &
	\includegraphics[scale=0.6]{gen-images/#1/#2_bg.png} &
	\includegraphics[scale=0.6]{gen-images/#1/#2_locations.png} \\
}

\begin{figure*}[t!]
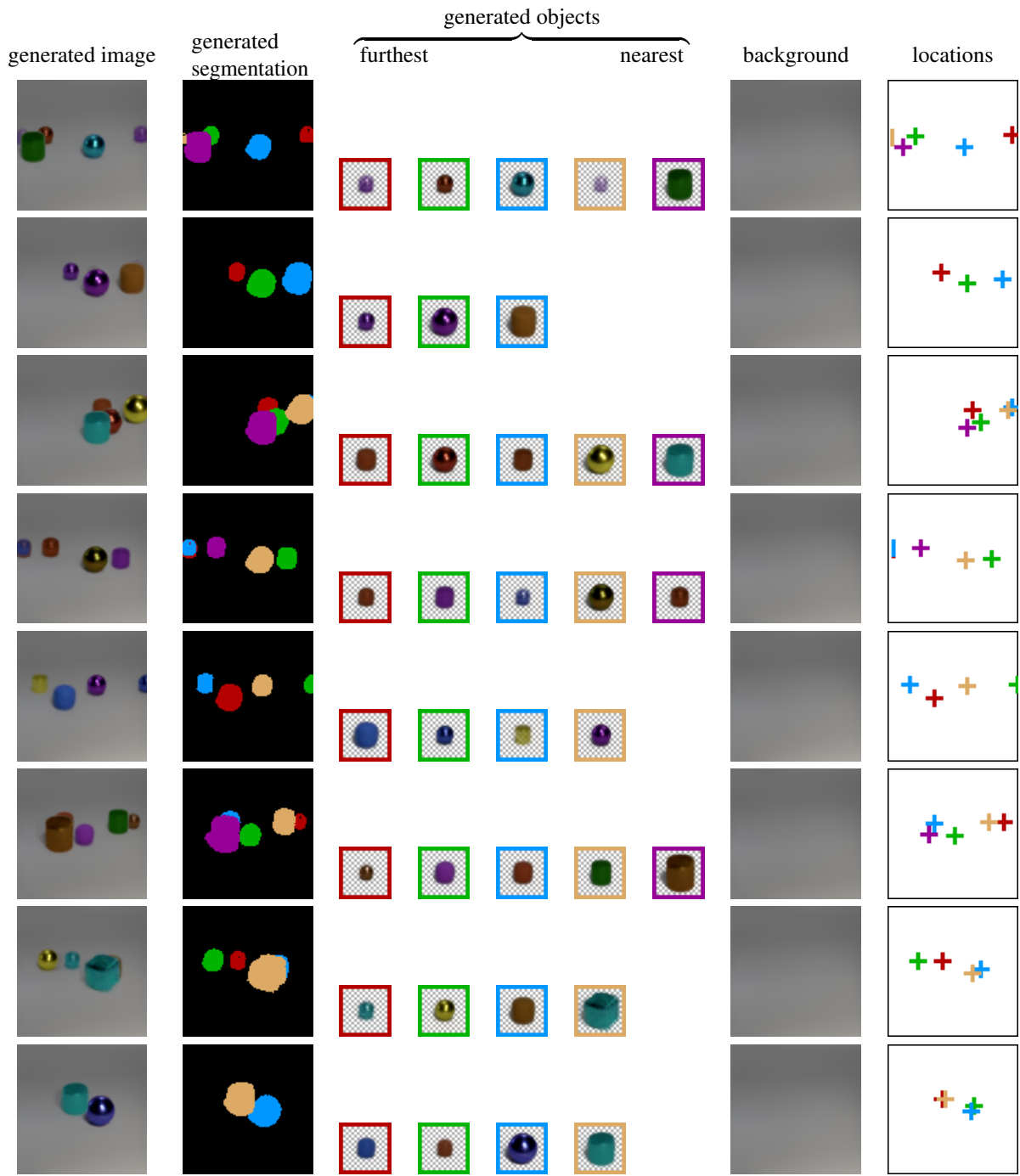

	\centering
	\begin{tabular}{@{}ccccccccc@{}}
		generated image & \parbox{5em}{generated \\ segmentation} & \multicolumn{5}{c}{
			$\overbrace{\text{~furthest}\hspace{3cm}\text{nearest~}}^{\mbox{\normalsize\text{generated objects}}}$
		} & background & locations \\
		\genrowclevrfive{1}{193}
		\genrowclevrfive{1}{216}
		\genrowclevrfive{1}{230}
		\genrowclevrfive{1}{233}
		\genrowclevrfive{1}{237}
		\genrowclevrfive{1}{420}
		\genrowclevrfive{1}{516}
		\genrowclevrfive{1}{554}
	\end{tabular}
	\\
	\caption{CLEVR-5 generations.}
	\label{fig:gen-examples-clevr5}
\end{figure*}

\begin{figure*}[t]
	\centering
	\begin{tabular}{@{}cccccccccc@{}}
		original image & reconstruction & \parbox{5em}{predicted \\ segmentation} & \multicolumn{5}{c}{
			$\overbrace{\text{~furthest}\hspace{3cm}\text{nearest~}}^{\mbox{\normalsize\text{reconstructed objects}}}$
		} & background & locations \\
		\reconrowclevrfive{7}{038}
		\reconrowclevrfive{7}{042}
		\reconrowclevrfive{7}{030}
		\reconrowclevrfive{7}{046}
		\reconrowclevrfive{7}{035}
		\reconrowclevrfive{7}{052}
		\reconrowclevrfive{7}{061}
		\reconrowclevrfive{7}{069}
	\end{tabular}
	\caption{CLEVR-5-vbg decompositions.}
	\label{fig:decomp-examples-clevr5vbg}
\end{figure*}

\begin{figure*}[t!]
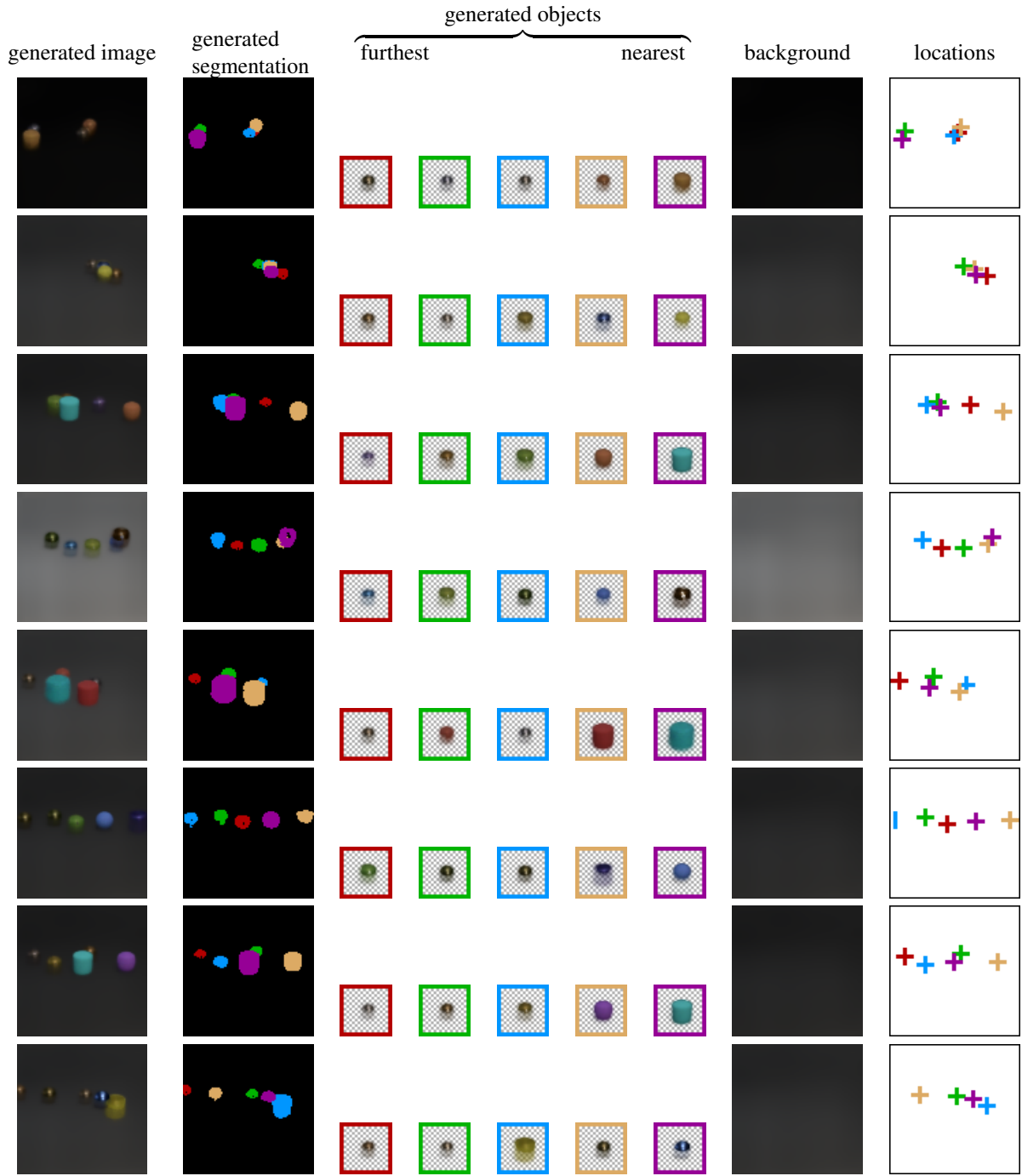

	\centering
	\begin{tabular}{@{}ccccccccc@{}}
		generated image & \parbox{5em}{generated \\ segmentation} & \multicolumn{5}{c}{
			$\overbrace{\text{~furthest}\hspace{3cm}\text{nearest~}}^{\mbox{\normalsize\text{generated objects}}}$
		} & background & locations \\
		\genrowclevrfive{7}{038}
		\genrowclevrfive{7}{042}
		\genrowclevrfive{7}{030}
		\genrowclevrfive{7}{046}
		\genrowclevrfive{7}{035}
		\genrowclevrfive{7}{052}
		\genrowclevrfive{7}{061}
		\genrowclevrfive{7}{069}
	\end{tabular}
	\\
	\caption{CLEVR-5-vbg generations. These samples show a failure case of $\alpha$ channels not always being binary. This issue can be resolved by adding a regularizer $\alpha \times \log(\alpha)$ to the final loss.}
	\label{fig:gen-examples-clevr5vbg}
\end{figure*}

\renewcommand{\reconrowclevrthree}[2]{ 
	\includegraphics[scale=0.7]{recon-images/reconstruction-#1/#2_original.png} &
	\includegraphics[scale=0.7]{recon-images/reconstruction-#1/#2_final.png} &
	\includegraphics[scale=0.7]{recon-images/reconstruction-#1/#2_segmentation.png} &
	\includegraphics[scale=0.7]{recon-images/reconstruction-#1/#2_obj_0.png} &
	\includegraphics[scale=0.7]{recon-images/reconstruction-#1/#2_obj_1.png} &
	\includegraphics[scale=0.7]{recon-images/reconstruction-#1/#2_obj_2.png} &
	\includegraphics[scale=0.7]{recon-images/reconstruction-#1/#2_bg.png} &
	\includegraphics[scale=0.7]{recon-images/reconstruction-#1/#2_locations.png}  \\
}

\begin{figure*}[t!]
	\centering
	\begin{tabular}{@{}cccccccc@{}}
		original image & reconstruction & \parbox{5em}{predicted \\ segmentation} & \multicolumn{3}{c}{
			$\overbrace{\text{~furthest}\hspace{1cm}\text{nearest~}}^{\mbox{\normalsize\text{reconstructed objects}}}$
		} & background & locations \\
		\reconrowclevrthree{0}{000}
		\reconrowclevrthree{0}{001}
		\reconrowclevrthree{0}{002}
		\reconrowclevrthree{0}{003}
		\reconrowclevrthree{0}{004}
		\reconrowclevrthree{0}{371}
		\reconrowclevrthree{0}{349}
		\reconrowclevrthree{0}{393}
	\end{tabular}
	\\
	\caption{CLEVR-3 decompositions.}
	\label{fig:rec-examplesclevr3}
\end{figure*}

\newcommand{\genrowclevrthree}[2]{ 
	\includegraphics[scale=0.7]{gen-images/#1/#2_final.png} &
	\includegraphics[scale=0.7]{gen-images/#1/#2_segmentation.png} &
	\includegraphics[scale=0.7]{gen-images/#1/#2_obj_0.png} &
	\includegraphics[scale=0.7]{gen-images/#1/#2_obj_1.png} &
	\includegraphics[scale=0.7]{gen-images/#1/#2_obj_2.png} &
	\includegraphics[scale=0.7]{gen-images/#1/#2_bg.png} &
	\includegraphics[scale=0.7]{gen-images/#1/#2_locations.png}  \\
}

\begin{figure*}[t!]
	\centering
	\begin{tabular}{@{}ccccccccc@{}}
		generated image & \parbox{5em}{generated \\ segmentation} & \multicolumn{3}{c}{
			$\overbrace{\text{~furthest}\hspace{1cm}\text{nearest~}}^{\mbox{\normalsize\text{generated objects}}}$
		} & background & locations \\
		\genrowclevrthree{0}{468}
		\genrowclevrthree{0}{566}
		\genrowclevrthree{0}{614}
		\genrowclevrthree{0}{618}
		\genrowclevrthree{0}{626}
		\genrowclevrthree{0}{627}
		\genrowclevrthree{0}{628}
		\genrowclevrthree{0}{633}
	\end{tabular}
	\\
	\caption{CLEVR-3 generations.}
	\label{fig:gen-examples-clevr3}
\end{figure*}

\begin{figure*}
	\centering

	\includegraphics[width=0.12\linewidth]{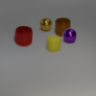}
	\includegraphics[width=0.12\linewidth]{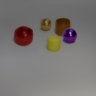}
	\includegraphics[width=0.12\linewidth]{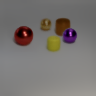}
	\includegraphics[width=0.12\linewidth]{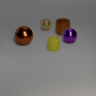}
	\includegraphics[width=0.12\linewidth]{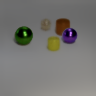}
	\includegraphics[width=0.12\linewidth]{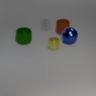}
	\\
	\includegraphics[width=0.12\linewidth]{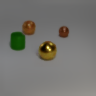}
	\includegraphics[width=0.12\linewidth]{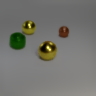}
	\includegraphics[width=0.12\linewidth]{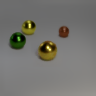}
	\includegraphics[width=0.12\linewidth]{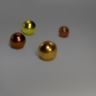}
	\includegraphics[width=0.12\linewidth]{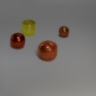}
	\includegraphics[width=0.12\linewidth]{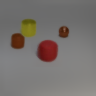}
	\\
	\includegraphics[width=0.12\linewidth]{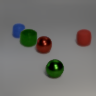}
	\includegraphics[width=0.12\linewidth]{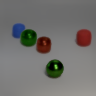}
	\includegraphics[width=0.12\linewidth]{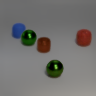}
	\includegraphics[width=0.12\linewidth]{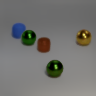}
	\includegraphics[width=0.12\linewidth]{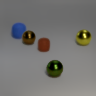}
	\includegraphics[width=0.12\linewidth]{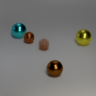}
	\\

    ~\\

	\includegraphics[width=0.12\linewidth]{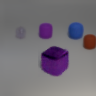}
	\includegraphics[width=0.12\linewidth]{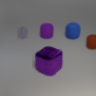}
	\includegraphics[width=0.12\linewidth]{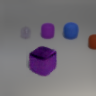}
	\includegraphics[width=0.12\linewidth]{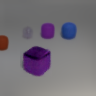}
	\includegraphics[width=0.12\linewidth]{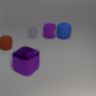}
	\includegraphics[width=0.12\linewidth]{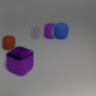}
	\\

	\includegraphics[width=0.12\linewidth]{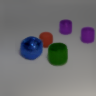}
	\includegraphics[width=0.12\linewidth]{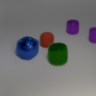}
	\includegraphics[width=0.12\linewidth]{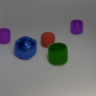}
	\includegraphics[width=0.12\linewidth]{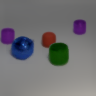}
	\includegraphics[width=0.12\linewidth]{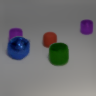}
	\includegraphics[width=0.12\linewidth]{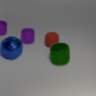}
	\\

	\includegraphics[width=0.12\linewidth]{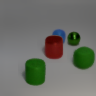}
	\includegraphics[width=0.12\linewidth]{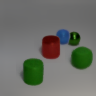}
	\includegraphics[width=0.12\linewidth]{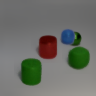}
	\includegraphics[width=0.12\linewidth]{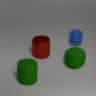}
	\includegraphics[width=0.12\linewidth]{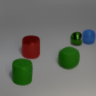}
	\includegraphics[width=0.12\linewidth]{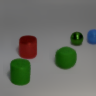}
	\\

    ~\\
	\includegraphics[width=0.12\linewidth]{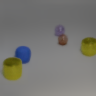}
	\includegraphics[width=0.12\linewidth]{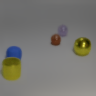}
	\includegraphics[width=0.12\linewidth]{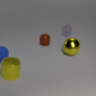}
	\includegraphics[width=0.12\linewidth]{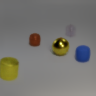}
	\includegraphics[width=0.12\linewidth]{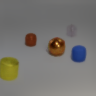}
	\includegraphics[width=0.12\linewidth]{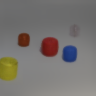}
	\\
	\includegraphics[width=0.12\linewidth]{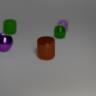}
	\includegraphics[width=0.12\linewidth]{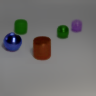}
	\includegraphics[width=0.12\linewidth]{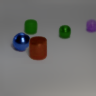}
	\includegraphics[width=0.12\linewidth]{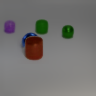}
	\includegraphics[width=0.12\linewidth]{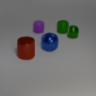}
	\includegraphics[width=0.12\linewidth]{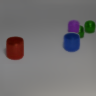}
	\\

	\includegraphics[width=0.12\linewidth]{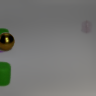}
	\includegraphics[width=0.12\linewidth]{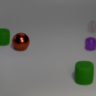}
	\includegraphics[width=0.12\linewidth]{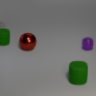}
	\includegraphics[width=0.12\linewidth]{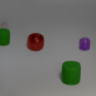}
	\includegraphics[width=0.12\linewidth]{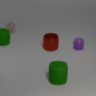}
	\includegraphics[width=0.12\linewidth]{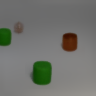}
	\\
	\caption{Disentangled interpolations in latent space. For each row, we interpolate between two images in latent space.
		\textbf{Top three rows}: we keep the object positions fixed as their appearances vary.
		\textbf{Middle three rows}: we vary the object positions, while keeping the appearance fixed.
        \textbf{Bottom three rows}: we allow all parameters to vary jointly.
		We see that the structured latent space allows disentangling of position and appearance variations.}
\end{figure*}

\end{document}